 \documentclass[final,5p,times,twocolumn,authoryear]{elsarticle}

\usepackage{amssymb}
\usepackage{lipsum}
\usepackage{graphicx}
\usepackage{amsmath}
\usepackage{hyperref}
\usepackage{bm}
\usepackage{blindtext}
\usepackage{xcolor}
\usepackage{subcaption}
\usepackage[percent]{overpic}

\journal{arXiv}

\begin{document}

\begin{frontmatter}



\title{A novel Fourier neural operator framework for classification of multi-sized images: Application to three dimensional digital porous media}


\author[first]{Ali Kashefi\corref{cor1}}
\ead{kashefi@stanford.edu}
\affiliation[first]{organization={Department of Civil and Environmental Engineering, Stanford University},
            city={Stanford},
            postcode={94305}, 
            state={CA},
            country={USA}}
\cortext[cor1]{Corresponding author.}


\author[third]{Tapan Mukerji}
\ead{mukerji@stanford.edu}
\affiliation[third]{organization={Department of Energy Science and Engineering, Stanford University},
            city={Stanford},
            postcode={94305}, 
            state={CA},
            country={USA}}

\begin{abstract}
Fourier neural operators (FNOs) are invariant with respect to the size of input images, and thus images with any size can be fed into FNO-based frameworks without any modification of network architectures, in contrast to traditional convolutional neural networks (CNNs). Leveraging the advantage of FNOs, we propose a novel deep-learning framework for classifying images with varying sizes. Particularly, we simultaneously train the proposed network on multi-sized images. As a practical application, we consider the problem of predicting the label (e.g., permeability) of three-dimensional digital porous media. To construct the framework, an intuitive approach is to connect FNO layers to a classifier using adaptive max pooling. First, we show that this approach is only effective for porous media with fixed sizes, whereas it fails for porous media of varying sizes. To overcome this limitation, we introduce our approach: instead of using adaptive max pooling, we use static max pooling with the size of channel width of FNO layers. Since the channel width of the FNO layers is independent of input image size, the introduced framework can handle multi-sized images during training. We show the effectiveness of the introduced framework and compare its performance with the intuitive approach through the example of the classification of three-dimensional digital porous media of varying sizes.
\end{abstract}




\end{frontmatter}




\section{Introduction and motivation}
\label{Sect1}

Since 2020, neural operators have gained extensive popularity, specifically with two versions of graph neural operators \citep{li2020neuralGraph,li2020multipole} and Fourier neural operators (FNOs) \citep{li2020fourierFNO,kovachki2024operator}. In this article, our attention is on FNOs. From a computer science perspective, regular FNOs fall in the category of supervised deep learning framework, necessitating a large volume of labeled data for training. FNOs have demonstrated their proficiency in input-output mapping across various industrial and scientific applications such as incompressible flows \citep{li2022fourier,bonev2023spherical,peng2024fourierCD,a16010024,MultiFidelity,gupta2022multispatiotemporalscale,attentionFNO}, wave equations \citep{zhu2023fourierBN,zou2023deepWave,RapidWave2}, thermal fields \citep{zhao2024recfno,hao2023gnot}, carbon storages and sequestration \citep{wen2022u,jiang2023fouriermionet}, and other areas \citep{peng2023rapid,you2022nonlocal,kontolati2023influence,zhu2023reliable,hua2023basis,white2023physics,li2021physicsA,pathak2022fourcastnet,rahman2022u,rahman2022generative,yang2022large,li2022fourierB,maust2022fourier,zhao2022incremental,renn2023forecasting,xiong2023koopmanlab,chen2023laplace,huang2023neuralstagger,poels2023fast,white2023speeding,thodi2023fourierTrafic,zhao2023local,tran2023factorized,lee2022meshindependent,brandstetter2023clifford,li2023geometryinformed,majumdar2023how,FNOclimate,lehmann20243d,subramanian2024towards,fanaskov2024spectral,lanthaler2021computation,azzizadenesheli2023neural}. From a computer vision perspective, these are framed as segmentation problems, where an input image, such as the geometry of an airfoil, is mapped to another image, for instance, the velocity field around that airfoil. An analogous area in computer vision is classification, where an input image is mapped, for example, to a name or number. While FNOs have potential in classification tasks, there exists only a limited number of research conducted in this application as per our knowledge \citep{ImageClassification1,RemoteSensingClassification,FNOclassification2}.

\cite{ImageClassification1} used the FNO architecture for classifying images in the CIFAR-10 dataset, containing ten different classes; however, they trained the network only on images with a fixed size of 32 by 32 pixels. Additionally, \cite{FNOclassification2} examined the FNO architecture for image classification. Although they tested images of various sizes (e.g., 28 by 28 pixels, 112 by 112 pixels, etc.), they trained and then tested the network separately for each size, assessing its performance on the corresponding size. \cite{RemoteSensingClassification} utilized the FNO architecture for the hyperspectral remote sensing image classification. Their dataset comprised images of various sizes, including 512 by 614 pixels, 610 by 340 pixels, and 512 by 217 pixels. However, they adjusted all images to a fixed size by adding patches. Consequently, although they employed the FNO architecture, in practice, they limited their analysis to images of a uniform size. In the current study, we narrow our focus on classification problems. More specifically, we consider the problem of predicting the permeability of three-dimensional digital porous media, which vary in size, as a benchmark test case.

\begin{figure*}
	\centering 
	\includegraphics[width=0.9\textwidth]{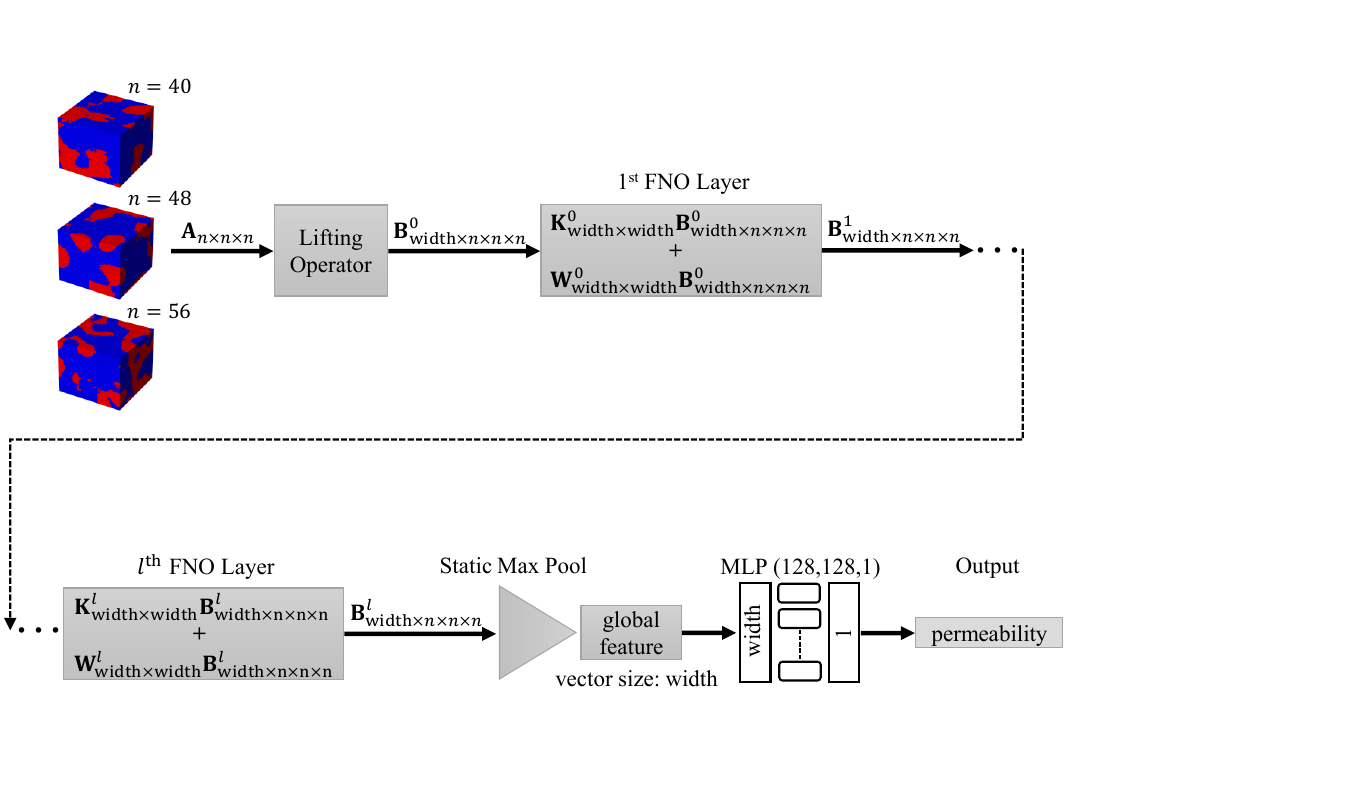}	
	\caption{Schematic of the proposed FNO-based framework for multi-size image classification} 
	\label{Fig1}
\end{figure*}

\begin{figure*}
	\centering 
	\includegraphics[width=0.9\textwidth]{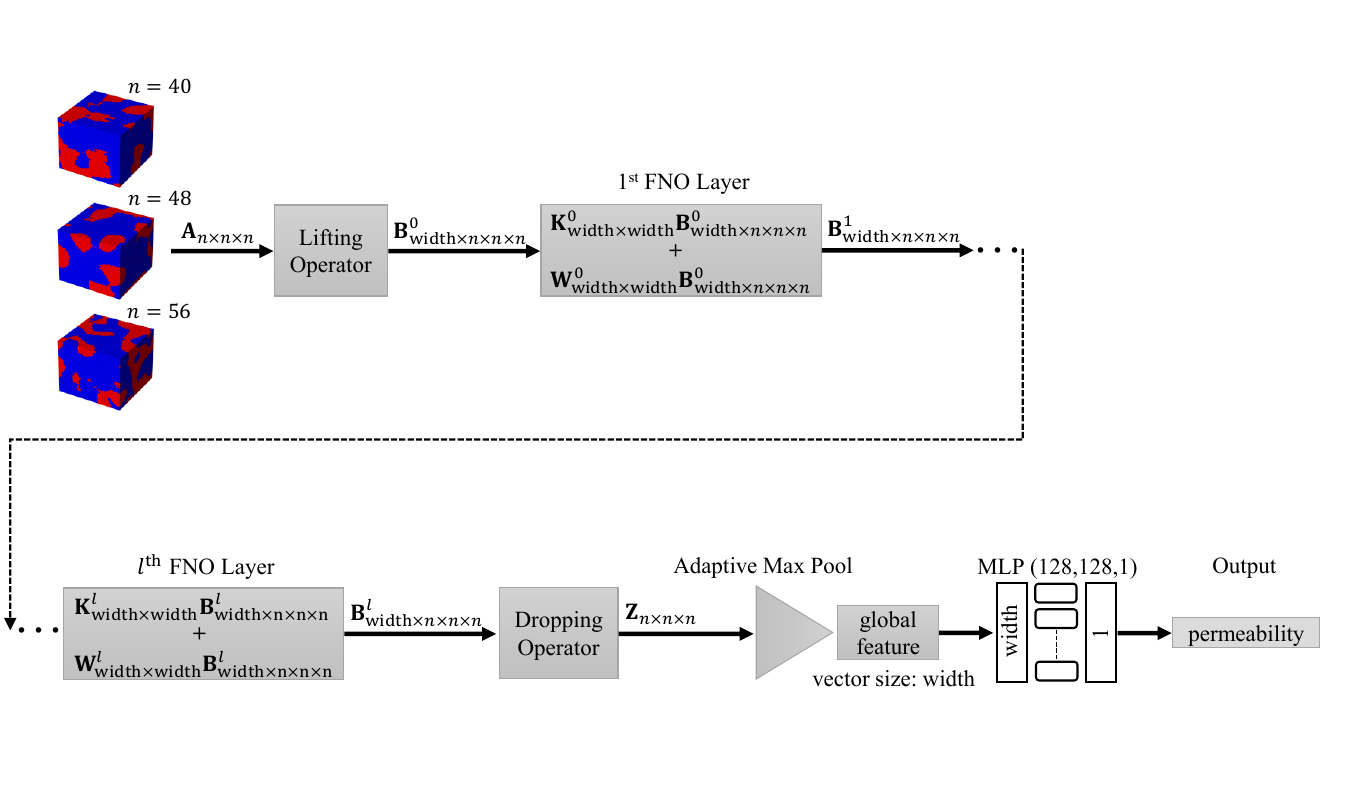}	
	\caption{Schematic of the intuitive FNO-based framework for multi-size image classification} 
	\label{Fig2}
\end{figure*}

FNOs are invariant with respect to the size of input images, and this characteristic ensures that images of varying sizes can be processed by FNO-based deep learning frameworks without requiring any architectural alterations. Note that regular convolutional neural networks (CNNs) lack this feature \citep{Goodfellow2016}. Building on this strength of FNOs, we introduce a deep-learning framework for training the network simultaneously on images with varying sizes for a classification problem. To achieve this deep learning framework, FNO layers must be connected to a classifier, which is commonly a multilayer perceptron (MLP). An intuitive approach to set this would be to link FNO layers to a classifier via adaptive max pooling. Considering the application of permeability prediction of three-dimensional porous media, our machine-learning experiments show that this intuitive approach only works well for porous media with fixed sizes. Pivoting from this, we propose our novel approach. Rather than using adaptive max pooling, we implement static max pooling with the size of the channel width of the FNO layers. Given that the size of the channel width of FNO layers is independent of the size of input images, our proposed framework can be efficiently trained on various image sizes at once (see Fig. \ref{Fig1} and Fig. \ref{Fig2}).

To explain, at a high level, the difference between using adaptive max pooling (see Fig. \ref{Fig2}) and static max pooling (see Fig. \ref{Fig1}), let us consider for example a three-dimensional image being fed as an input of the deep learning framework. For both pooling methods, at the framework's outset, FNO layers lift the input image from its three-dimensional space to a higher dimensional space, determined by the size of the channel width of the FNO layers. In the case of adaptive max pooling, after FNO layer operations, the outcome eventually is dropped into the initial three-dimensional space with the same size as the input image. This array then serves as the input of adaptive max pooling. The output of the adaptive pooling is then the input of the classifier. In the case of static max pooling, before FNO layers drop the output, we implement static max pooling, which functions within the high dimensional space and pools with the size of the channel width of FNO layers. The resulting output from this pooling then becomes the classifier's input. A more detailed exploration of these concepts is provided in Sect. \ref{Sect2}.

The study of physical and geometric features of porous media is important in diverse scientific and industrial areas such as digital rock physics \citep{ANDRApart1,ANDRApart2}, membrane systems \citep{LIANG2023116359}, geological carbon storages \citep{BLUNT2013197}, and medicine \citep{PorousMedicine,Das2018}. Deep learning frameworks have been widely used for predicting the permeability of porous media \citep{meng2023transformer,PermCurve,kashefi2023PorousMediaPIPN,kashefi2021PointNetPorousMedia,Mingliang1,Hong2020RapidEstimation,WU20181215,Tembely2020DeepLearningPermeability,MASROOR2023,consideringorganicmatter}, but, to the best of our knowledge, all these frameworks were trained on a fixed-size porous media. Note that training the proposed network to predict the permeability of porous media of varying sizes comes with an exclusive challenge when compared to training the network on conventional images for the purpose of classifying them by their names (like those of cats and dogs). For conventional images, one possible solution to handle images with different sizes is to equalize them by adding mini patches to the smaller images. Nevertheless, this solution is inapplicable to the porous media problem. Adding mini patches to porous media can alter their physical properties such as permeability. For instance, adding mini patches around a porous medium simulates sealing it with wall boundaries, which prohibits flow within its pore spaces, resulting in a permeability of zero. Additionally, the inherently three-dimensional nature of porous media introduces another layer of complexity compared to the two-dimensional conventional images. We summarize the contributions of our work in the following bullet points:

\begin{itemize}

    \item We propose a novel deep-learning framework for image classification.

    \item The proposed framework leverages Fourier neural operators, which are invariant to the size of input images.

    \item Specifically designed to train simultaneously on images of multiple sizes, the framework can effectively classify images of varying sizes.

    \item This is an important feature for applications where input images naturally vary in size. We demonstrate its application specifically for three-dimensional images.
\end{itemize}

The remainder of this article is organized as follows. We introduce and discuss the concept of Fourier neural operators for image classification in Sect. \ref{Sect2}, starting with the traditional strategy of adaptive max pooling, followed by our novel approach of static max pooling in the high dimension of the Fourier space channel. A brief review of theoretical aspects of FNOs is given in Sect. \ref{Sect23}. Data generation and the training methodologies are respectively presented in Sect. \ref{Sect3} and Sect. \ref{Sect4}. In Sect. \ref{Sect5}, we provide results and discussion, including a comparison between traditional strategy and our novel approach. Moreover, we present a sensitivity analysis, covering the number of Fourier modes, the channel width of discrete Fourier space, the number of FNO units, and the effect of activation functions and average pooling. The deep learning model generalizability is discussed in this section as well. Finally, we summarize the work and present insight into future directions in Sect. \ref{Sect6}.


\section{Fourier neural operators for image classification}
\label{Sect2}
\subsection{Our novel approach: Static max pooling in channel width of FNO layers}
\label{Sect21}

In this subsection, we introduce the architecture of our proposed deep learning framework. Our explanation heavily uses matrix notation to ensure clarity and provide a deeper understanding. As illustrated in Fig. \ref{Fig1}, the input of the deep learning framework is a cubic binary porous medium represented as the matrix $\mathbf{A}_{n \times n \times n}$. As a first step, the matrix $\mathbf{A}_{n \times n \times n}$ is lifted to a higher dimensional space using a fully connected network. The dimension of this space is termed the channel width of an FNO layer, shown by ``width'' in our matrix notation. This lifting results in a four-dimensional matrix, denoted as $\mathbf{B}_{\text{width} \times n \times n \times n}^0
$. The matrix $\mathbf{B}_{\text{width} \times n \times n \times n}^0
$ becomes subsequently the input of an FNO layer. Within the FNO layer, two operations are applied to $\mathbf{B}_{\text{width} \times n \times n \times n}^0
$: the kernel integration operator, denoted by $\mathbf{K}_{\text{width} \times \text{width}}^0
$, and the linear transformation operator, denoted by $\mathbf{W}_{\text{width} \times \text{width}}^0$. The network computes the matrix-matrix multiplication of $\mathbf{K}_{\text{width} \times \text{width}}^0\mathbf{B}_{\text{width} \times n \times n \times n}^0
$ and $\mathbf{W}_{\text{width} \times \text{width}}^0 \mathbf{B}_{\text{width} \times n \times n \times n}^0$ and then sums up the resulting matrices, as depicted in Fig. \ref{Fig1}. The output undergoes element-wise operations of the Rectified Linear Unit (ReLU) activation function \citep{Goodfellow2016} defined as

\begin{equation}
    \sigma(\gamma) = \max(0, \gamma),
    \label{Eq1}
\end{equation}
Resulting in a four-dimensional matrix $\mathbf{B}_{\text{width} \times n \times n \times n}^1$. Mathematically, this procedure can be summarized as

\begin{align}
    \mathbf{B}_{\text{width} \times n \times n \times n}^1 &= \sigma\left(\mathbf{K}_{\text{width} \times \text{width}}^0 \mathbf{B}_{\text{width} \times n \times n \times n}^0 \right. \nonumber \\
    &\quad \left. + \mathbf{W}_{\text{width} \times \text{width}}^0 \mathbf{B}_{\text{width} \times n \times n \times n}^0 \right).
    \label{Eq2}
\end{align}
In scenarios where multiple FNO layers exist in the framework, the matrix $\mathbf{B}_{\text{width} \times n \times n \times n}^1$ serves as the input for the succeeding FNO layers, and the same sequence of operations is applied. If we assume that there are $l$ number of FNO layers, the output from the final FNO layer is the matrix $\mathbf{B}_{\text{width} \times n \times n \times n}^l$. In the next step, we implement static max pooling on the first dimension of matrix $\mathbf{B}_{\text{width} \times n \times n \times n}^l$. Because ``width'' is independent of the input image dimension (i.e., $n$), the static pooling works for input images with any desired size (e.g., $n=40$, $n=48$, and $n=56$). Note that ``width'' is a hyper parameter of FNO layers and independent of $n$, as all the matrix-matrix multiplication operates on the dimension with the size ``width'', and not ``$n$''. The static max pooling produces a vector of length width, representing the global features of the input images. The vector is then connected to a classifier. The classifier is a Multilayer Perceptron (MLP) composed of three layers of sizes 128, 128, and 1. The ReLU activation function is used in the initial two layers along with a dropout with a rate of 0.3. Following the third layer, a sigmoid activation function defined as 

\begin{equation}
    \sigma(\gamma) = \frac{1}{1 + e^{-\gamma}},
    \label{Eq3}
\end{equation}
is used to ensure output values are bounded between 0 and 1.

\subsection{Intuitive approach: Adaptive max pooling in 3D spatial space}
\label{Sect22}

In this subsection, we explain the intuitive approach (see Fig. \ref{Fig2}). Drawing parallels to our approach elaborated in the previous subsection, we begin by considering the input porous medium, which is a three-dimensional matrix represented by $\mathbf{A}_{n \times n \times n}$. All operations outlined in Sect. \ref{Sect21} are applied to $\mathbf{A}_{n \times n \times n}$ until the network obtains the matrix $\mathbf{B}_{\text{width} \times n \times n \times n}^l$ at an intermediate step, as depicted in Fig. \ref{Fig2}. As the next step, we drop (as an inverse of the lifting operator explained in Sect. \ref{Sect21}) the matrix $\mathbf{B}_{\text{width} \times n \times n \times n}^l$ from the high dimensional space to the default space by means of a fully connected network. This transformation results in the matrix $\mathbf{Z}_{n \times n \times n}$. At this juncture, we use adaptive three-dimensional max pooling, a functionality that is available in deep learning platforms such as PyTorch \citep{paszke2019pytorch} or TensorFlow \citep{tensorflow2015}. To ensure a fair comparison between the traditional approach and our novel approach, we keep the size of the vector of the global feature consistent across both approaches. To this end, the output of the adaptive max pooling is tailored to yield a vector of size ``width''. The resulting vector represents the global features of the input images.

Note that because the size of matrix $\mathbf{Z}_{n \times n \times n}$ depends on the size of the input image (i.e., $n$), the pooling must be adaptive as we plan to train the network simultaneously on input images with varying sizes (e.g., $\mathbf{A}_{40 \times 40 \times 40}$,  $\mathbf{A}_{48 \times 48 \times 48}$, and $\mathbf{A}_{56 \times 56 \times 56}$). Subsequent to the adaptive max pooling, the global feature vector is connected to a classifier. This classifier features and architecture is precisely the same as the one elucidated in Sect. \ref{Sect21}.

To close this subsection, it is noted that the main difference between static max pooling and adaptive max pooling can be articulated as follows. In static max pooling, the kernel size and stride are constant, whereas in adaptive max pooling, they are not constant and are computed based on the input size. For further details and formulations, one may refer to the TensorFlow \citep{tensorflow2015} and PyTorch \citep{paszke2019pytorch} handbooks.

\subsection{A brief review of theoretical aspects of Fourier neural operators}
\label{Sect23}

We focused on the technical aspects and computer implementation of FNO layers in Sect. \ref{Sect21} and Sect. \ref{Sect22}. Theoretical aspects of FNO layers have already been vastly explained and discussed in the literature \citep{li2020fourierFNO}. In this subsection, we briefly review the theory behind FNO layers and highlight some important features.

As discussed in Sect. \ref{Sect21}, an FNO layer comprises two main operators: the integral kernel operator and the linear transformation. We overview the integral kernel operator. We consider the bounded domain $D$ such that $D \subset \mathbb{R}^d$, where $d$ indicates the physical dimensionality of the problem and is equal to 3 (i.e., $d=3$) for the current problem since we deal with three-dimensional porous media. We further show the input of the FNO layer by $b(x)$ with $x \in D$, where $b$ is a function representing all the operators applied to $x$ when it arrives at the gate of the FNO layer. Moreover, we define the periodic kernel function $\tau : \mathbb{R}^{2(d + d_a)} \rightarrow \mathbb{R}^{\text{width} \times \text{width}}
$, where $d_a$ is the number of input features and is equal to 1 (i.e., $d_a=1$) in this study, because the input of the deep learning framework is only a cubic binary array (representing a porous medium), and this array only provides one feature, which is the geometry of the porous medium. Additionally, recall that ``width'' is the channel width of the FNO layer, as illustrated in Sect. \ref{Sect21} and Sect. \ref{Sect22}. Following the formulation proposed by \cite{li2020fourierFNO}, the operation of the integral kernel $\mathcal{K}$ on the function $b(x)$ in the continuous space is defined as

\begin{equation}
    \mathcal{K}b(x) = \int_{D} \tau(x,y) b(y) \, dy, \quad \forall x \in D.
    \label{Eq4}
\end{equation}
Following the original design of FNO layers by \cite{li2020fourierFNO}, we introduce the condition $\tau(x, y) = \tau(x - y)$. By applying the convolution theorem as detailed in the literature \citep{li2020fourierFNO}, the following expression for the integral kernel operator is obtained:

\begin{equation}
\mathcal{K}b(x) = \mathcal{F}^{-1} \left( \mathcal{F}(\tau) \cdot \mathcal{F}(b(x)) \right), \quad \forall x \in D,
\label{Eq5}
\end{equation}
where the Fourier transform and its inverse are shown by $\mathcal{F}$ and $\mathcal{F}^{-1}$, respectively. We introduce $\mathcal{R}$ as the learnable Fourier transform of $\tau$ such that

\begin{equation}
    \mathcal{R} = \mathcal{F}(\tau).
    \label{Eq6}
\end{equation}
Beyond the theory, we must implement these mathematical concepts in a deep learning framework. In this way, we work with discrete spaces and consequently, discrete modes of the Fourier transform. Hence, $\mathcal{R}$ is implemented as a neural network. Additionally, each porous medium is represented by $n^3$ discrete points such that $\{x_1, x_2, \cdots, x_{n^3} \} \subset D
$. Moreover, Fourier series expansions are truncated at a maximum number of modes $m_\text{max}$ computed as

\begin{equation}
    m_{\text{max}} = \left| \left\{ m \in \mathbb{Z}^d : |m_j| \leq m_{\text{max}, j}, \text{ for } j = 1, \cdots, d \right\} \right|,
    \label{Eq7}
\end{equation}
where $m_{\text{max}, j}$ is the maximum number of modes taken in the dimension $j$, and is a hyper-parameter of the FNO layer. Note that since we work on three-dimensional problems in the current study, $d=3$, and thus, there are only $m_{\text{max},1}$, $m_{\text{max},2}$, and $m_{\text{max},3}$. As a result, the components of the $\mathcal{R} \cdot \mathcal{F}(b(x))$ operation can be computed by the following formulation

\begin{align}
    [\mathcal{R} \cdot \mathcal{F}(b(x))]_{m,i} &= \sum_{j=1}^{\text{width}} [\mathcal{R}]_{m,i,j} [\mathcal{F}(b(x))]_{m,j}, \\
    &\quad  m = 1, \cdots, m_{\text{max}}, \quad i = 1, \cdots, \text{width}, \nonumber
    \label{Eq8}
\end{align}
where $[\mathcal{R}] \in \mathbb{C}^{m_{\text{max}} \times \text{width} \times \text{width}}$ is the matrix representation of $\mathcal{R}$ in the discrete space. $[\mathcal{R} \cdot \mathcal{F}(b(x))] \in \mathbb{C}^{m_{\text{max}} \times \text{width}}
$ and $[\mathcal{F}(b(x))] \in \mathbb{C}^{m_{\text{max}} \times \text{width}}
$ are similarly defined. To increase computing efficiency and enable parallel computing, the operator $[\mathcal{R}]$, for the current three-dimensional problem, is better to be implemented as a five-dimensional matrix expressed as

\begin{equation}
    \mathbf{R}_{m_{\text{max},1} \times m_{\text{max},2} \times m_{\text{max},3} \times \text{width} \times \text{width}}.
    \label{Eq9}
\end{equation}
As can be seen from Eq. \ref{Eq9}, the size of matrix $\mathbf{R}$, and thus the count of trainable parameters in the FNO layer, is a function of the number of maximum Fourier modes at each dimension and the channel width of the FNO layer. Recall that these parameters (i.e., $m_{\text{max},1}$, $m_{\text{max},2}$, $m_{\text{max},3}$, and width) are the hyperparameter of FNO layers and need to be tuned by potential users.


\begin{figure*}[htp]
    \centering
    \begin{subfigure}{0.31\textwidth}
        \centering
        \begin{overpic}[width=\textwidth]{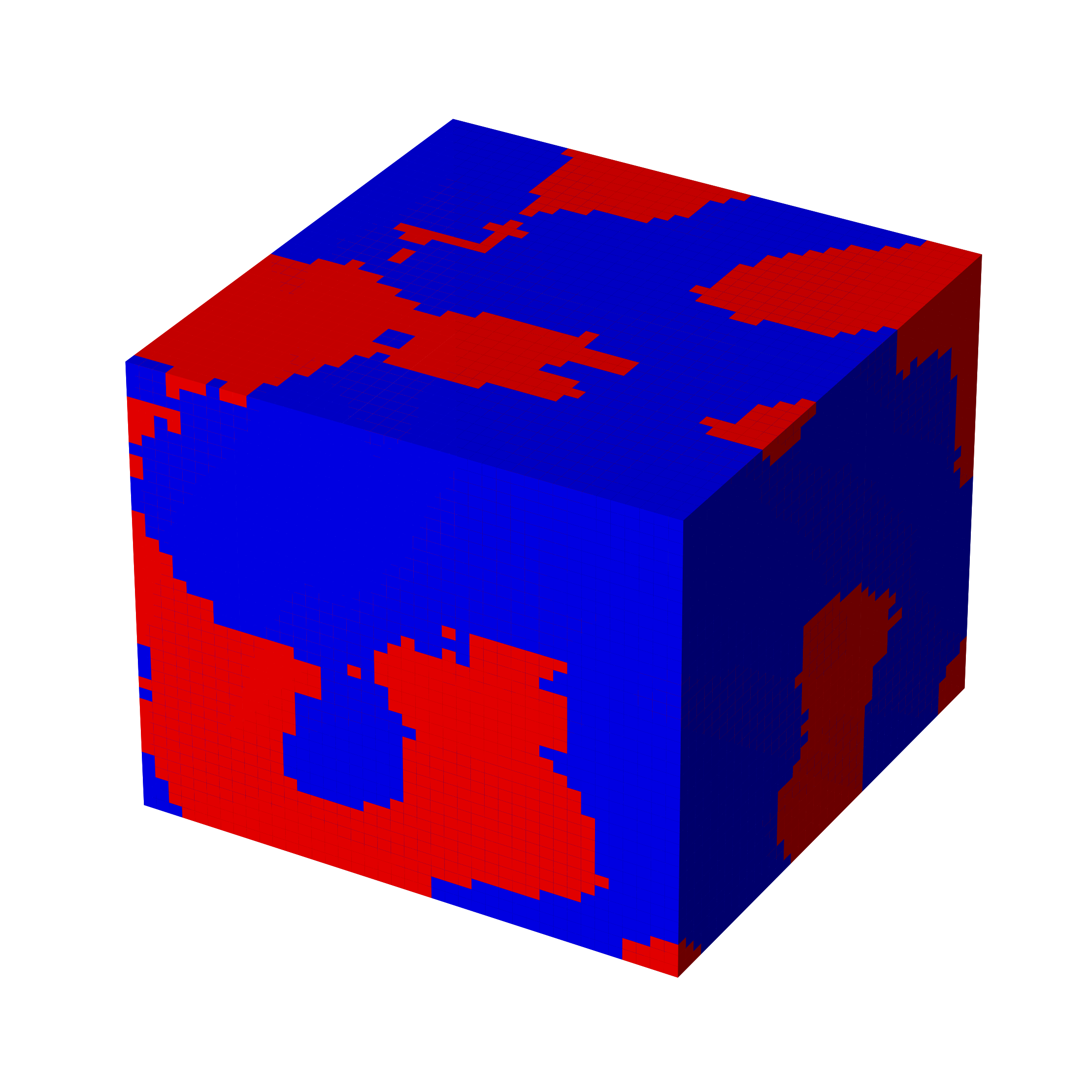}
            \put(5,90){\bfseries a}
        \end{overpic}
    \end{subfigure}
    \hfill
    \begin{subfigure}{0.31\textwidth}
        \centering
        \begin{overpic}[width=\textwidth]{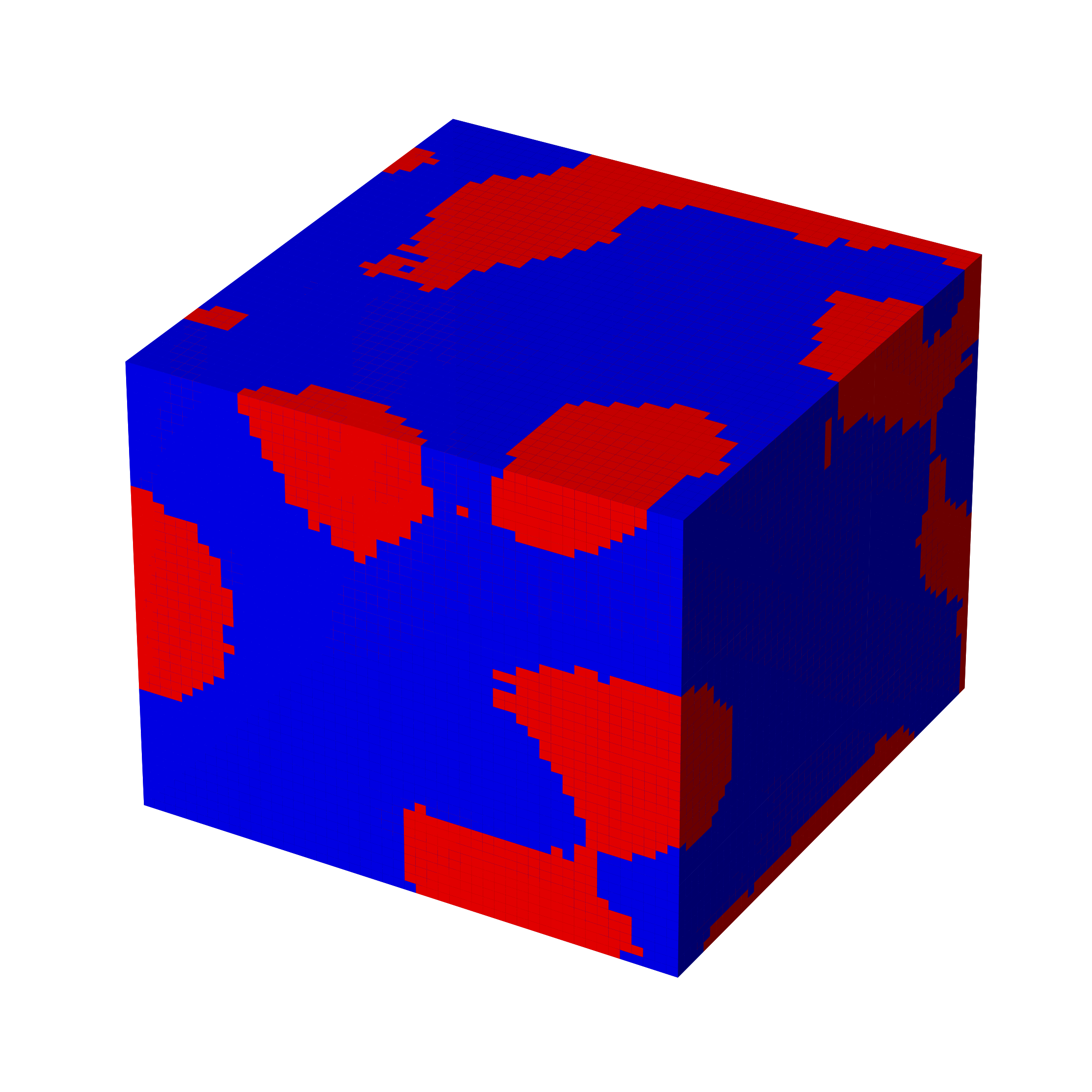}
            \put(5,90){\bfseries b}
        \end{overpic}
    \end{subfigure}
    \hfill
    \begin{subfigure}{0.31\textwidth}
        \centering
        \begin{overpic}[width=\textwidth]{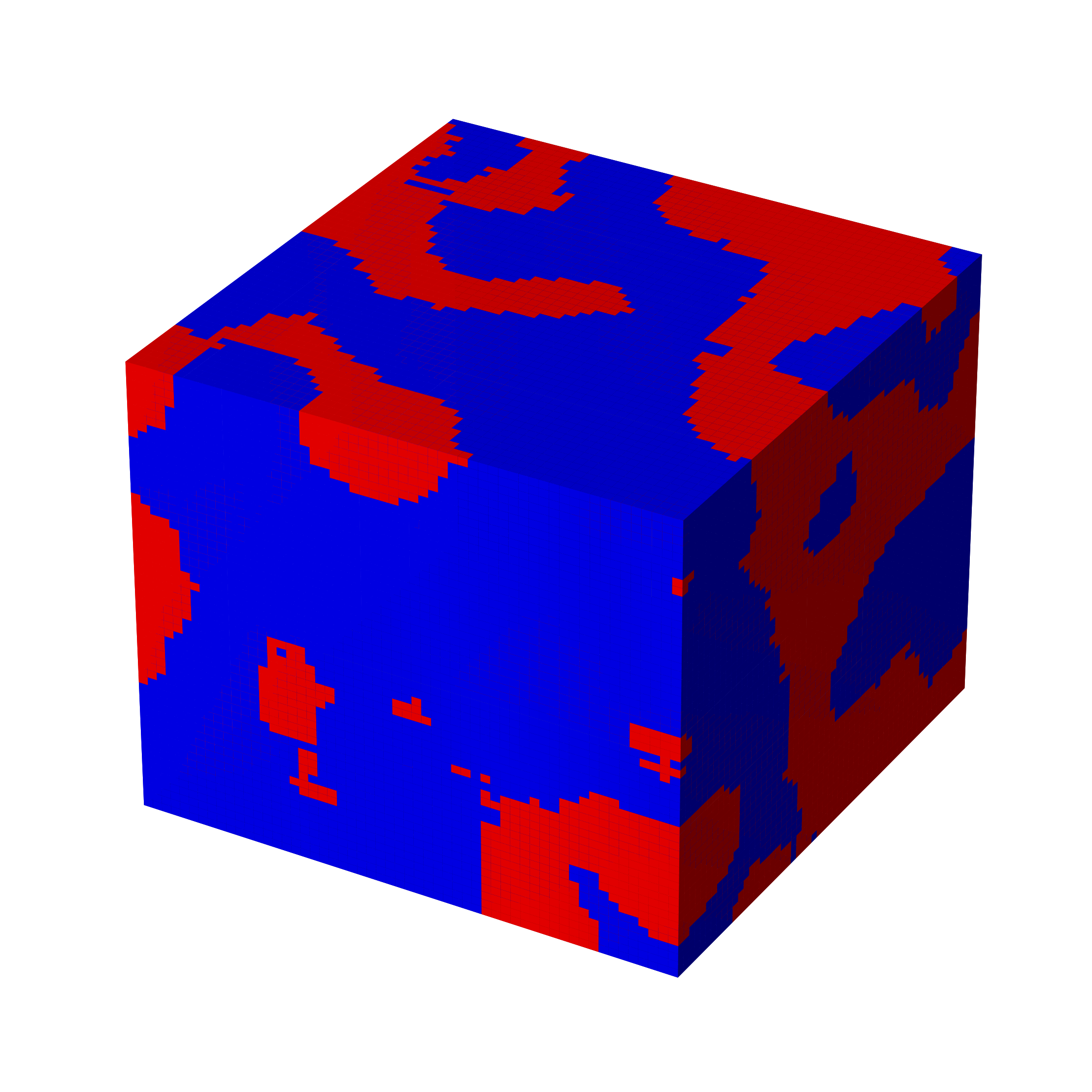}
            \put(5,90){\bfseries c}
        \end{overpic}
    \end{subfigure}
    \caption{A few examples of synthetically generated three-dimensional digital porous media for training the proposed neural network; \textbf{a} an image of size $40^3$, \textbf{b} an image of size $48^3$, and \textbf{c} an image of size $56^3$. Blue represents grain space, while red indicates pore space.}
    \label{Fig3}
\end{figure*}

\section{Data generation}
\label{Sect3}

To generate synthetic data to examine the deep learning framework under investigation in this study, we consider cubic porous medium domains with length L along each side, spatial correlation length of $l_c$, and porosity of $\phi$ (the ratio of pore spaces to the total volume of a porous medium). We use the truncated Gaussian algorithm \citep{Lantuejoul2002Geostatistical,LeRavalec2004Conditioning} to generate synthetic porous media. In practice, we create three-dimensional cubic arrays of dimension $n \times n \times n$, populated with random numbers conforming to a normal distribution with the characteristics of a mean value of 0.0 and a standard deviation of 1.0. Subsequently, we filter the arrays by a three-dimensional Gaussian smoothing kernel with a standard deviation of 5.0 and a filter size commensurate with a spatial correlation length ($l_c$) of 17. We then subject the arrays to a binarization process via a thresholding number such that the porosity ($\phi$) of the resulting arrays lies within the range of [0.125, 0.200]. We use the MATLAB software to handle the above-described steps. We set $L$ as $n \times \delta x$, where $\delta x$ represents the length of each side of a pixel in porous media. We set $\delta x$ to 0.003 m. We generate porous media with three different sizes by considering three different values for $n$, such that $n_1=40$, $n_2=48$, and $n_3=56$. In this way, each cubic porous medium can be characterized by its size as $n^3$ (e.g., $40^3$, $48^3$, and $56^3$). For each $n$, we generate 1250 data. We randomly split the generated data corresponding to each size into three categories of training (80\%, i.e., 1000 data), validation (10\%, i.e., 125 data), and test (10\%, i.e., 125 data). Hence, there are 3750 data in total, 3000 data for the training set, 375 data for the validation set, and 375 data for the test set. Figure \ref{Fig3} exhibits a few examples of the generated synthetic data.

To stimulate the incompressible viscous Newtonian flow within the generated porous media, we apply a constant pressure gradient in the $x$ direction ($\Delta p/L$). Zero velocity boundary condition is applied at the top and bottom of the porous medium on the $y-z$ planes. Given the geometry and boundary conditions illustrated above, we use a Lattice Boltzmann solver \citep{LBM} to solve the continuity and steady-state Stokes equations, which are written as follows:

\begin{equation}
    \nabla \cdot \bm{\mathit{u}}= 0, \quad \text{in } V,
    \label{Eq10}
\end{equation}

\begin{equation}
    -\nabla p + \mu \Delta \bm{\mathit{u}} = \textbf{0}, \quad \text{in } V,
    \label{Eq11}
\end{equation}
where $\mu$ is the dynamic viscosity, $\bm{\mathit{u}}$ and $p$ indicate, respectively, the velocity vector and pressure fields in the pore space of the porous medium, $V$. In the next step, we compute the permeability in the $x-$direction ($k$) using Darcy’s law \citep{Darcy1856Fontaines},

\begin{equation}
    k = -\frac{\mu \bar{U}}{\Delta p / L},
    \label{Eq12}
\end{equation}
where $\bar{U}$ shows the average velocity in the entire porous medium (i.e., including solid matrices). The computed permeabilities of our data set fall in the range [20 mD, 200 mD].

\section{Training}
\label{Sect4}
To accelerate the convergence of the training procedure, the output training data (i.e., permeability) are scaled in the range of [0, 1] using the maximum and minimum values of the training set. Note that although we train a single neural network simultaneously on porous media with three different sizes (corresponding to $n_1$, $n_2$, and $n_3$), we normalize the permeability of porous media of each size using the maximum and minimum values of the specific size. Mathematically, it can be written as

\begin{equation}
    \{\hat{k}_{\text{truth}}\}_{n_j} = \frac{\{k\}_{n_j} - \min \{k\}_{n_j}}{\max \{k\}_{n_j} - \min \{k\}_{n_j}}, \quad j = 1, 2, \text{and } 3,
    \label{Eq13}
\end{equation}
where, $\hat{k}_{\text{truth}}$ shows the ground truth scaled permeability. Moreover, for instance, $\{k\}_{n_1}$ indicates the training data containing porous media with the size of $40^3$ (because $n_1=40$). Note that we eventually rescale predicted permeability ($\hat{k}_{\text{prediction}}$) to the physical domain ($k_{\text{prediction}}$) for analyzing the neural network performances. Note that in the application of predicting the permeability of porous media using deep learning models, the presence of noisy data or outliers in the training set indicates that at least one sample of porous media has a permeability deviating significantly from the distribution observed in the rest of the training set. This means that data normalization, using Eq. \ref{Eq13}, would result in the permeability values of the training set clustering near 0 or 1. Such a scenario would seriously impair the training process of any neural network, including the one proposed in this study. Therefore, data cleaning is a crucial step before proceeding with data normalization. Concerning the loss function, we use the mean squared error function defined as

\begin{equation}
    \text{Loss} = \frac{1}{N} \sum_{i=1}^{N} (\hat{k}_{\text{prediction}} - \hat{k}_{\text{truth}})^2,
    \label{Eq14}
\end{equation}
where $N$ is the number of data in the training set (i.e., $N=3000$). Note that using relative mean squared error as the loss function does not lead to a significant difference in the results, based on our experiments. We set the number of modes in each dimension to 2 (i.e., set $m_{\text{max},1}=2$, $m_{\text{max},2}=2$, and $m_{\text{max},3}=2$). The channel width of the discrete Fourier space is set to 64 (i.e., $\text{width}=64$). It is worth noting that both the number of modes and the channel width play pivotal roles in the network performance. Detailed discussions on their significance and implications are provided in Sect. \ref{Sect522} and \ref{Sect523}, respectively. Additionally, we implement three units of FNOs in the network. The Adam optimizer \citep{kingma2014adam} is used. A constant learning rate of 0.001 is selected. We use the stochastic gradient descent \citep{Goodfellow2016} with a mini-batch size of 50. As discussed in Sect. \ref{Sect2}, the architecture of FNOs is designed to be independent of the spatial resolution of input images. During the training process, however, all the input images within a mini-batch must be the same size. In practice, each epoch of training is characterized by an inner loop that iterates through mini-batches of differing porous medium sizes (i.e., $40^3$, $48^3$, and $56^3$). Within this loop, the training process starts with a mini-batch of data of size $40^3$, followed by one of size $48^3$, and then continues to $56^3$, in sequence until all the data in the training set are covered within the epoch. Note that the trainable parameters of the network are updated only at the end of each epoch. Our deep learning experiments show that the order in which these differently sized porous media are fed within an epoch has no significant influence on the result accuracy and convergence speed, whether starting with the porous media of size $40^3$, followed by $48^3$ and $56^3$, or any other permutation. From a software perspective, we employ the NVIDIA A100 (SXM4) graphic card with 80 Gigabytes of RAM for training the networks.

In the last paragraph of this subsection, we address the metric used for assessing the effectiveness of permeability prediction. We use the coefficient of determination, also known as the $R^2$ score, which can be calculated using the following formula

\begin{equation}
    R^2 = 1 - \frac{\sum_{i=1}^{Q} ({k_{\text{truth}}}_i - {k_{\text{prediction}}}_i)^2}{\sum_{i=1}^{Q} ({k_{\text{truth}}}_i - \bar{k})^2},
    \label{Eq15}
\end{equation}
where $Q$ represents the number of the data in a set (e.g., training, test, etc.) and $\bar{k}$ is the average value of the set $\{{k_{\text{truth}}}_i\}_{i=1}^Q$.

\begin{figure}
	\centering 
	\includegraphics[width=0.4\textwidth]{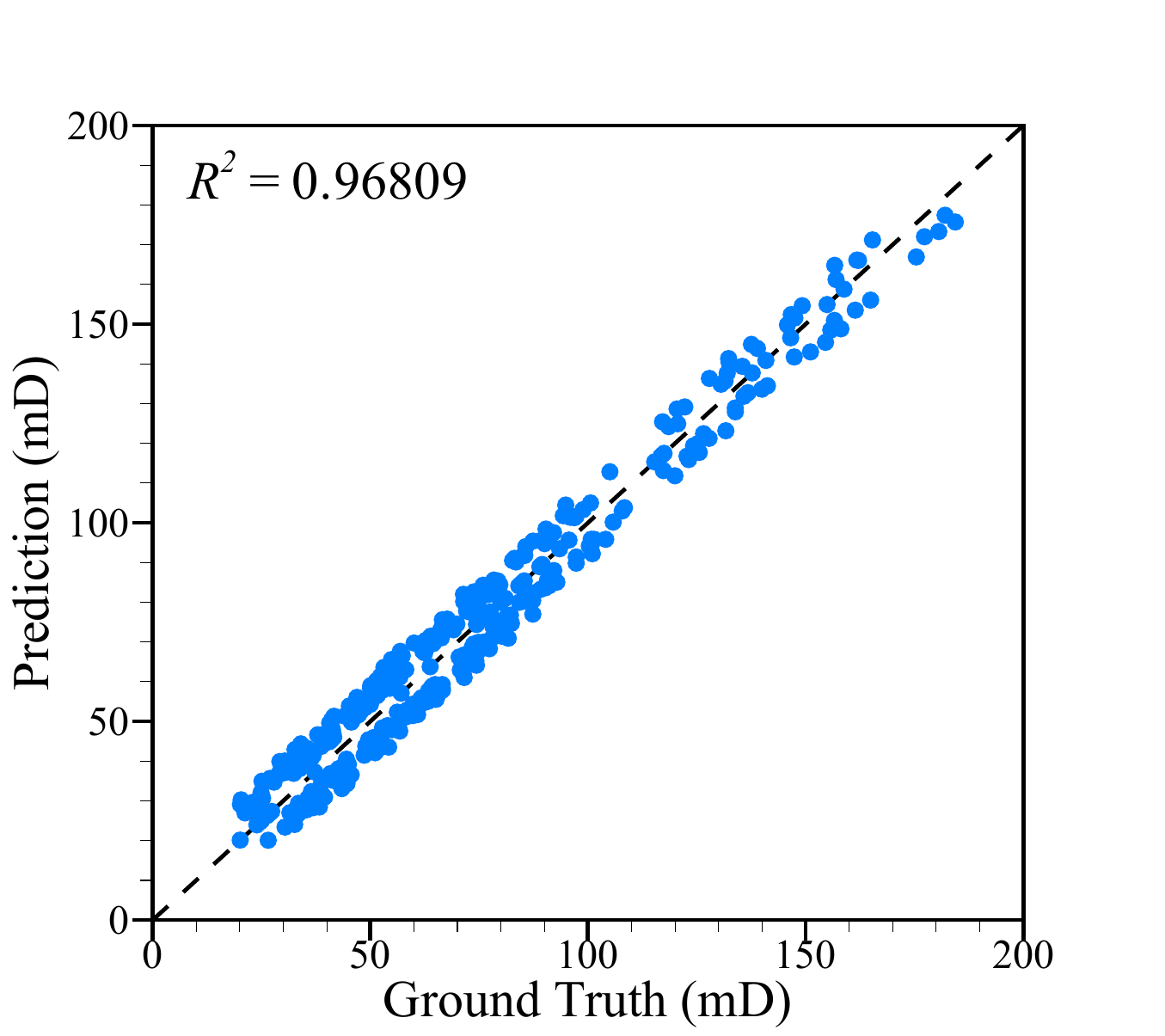}
    \caption{$R^2$ plots for the test set (375 data) using the proposed approach for classification of multi-sized images} 
	\label{Fig41}
\end{figure}

\section{Results and discussion}
\label{Sect5}

\subsection{General analysis}
\label{Sect521}

\begin{figure*}
    \centering
    \begin{subfigure}{0.33\textwidth}
        \centering
        \begin{overpic}[width=\textwidth]{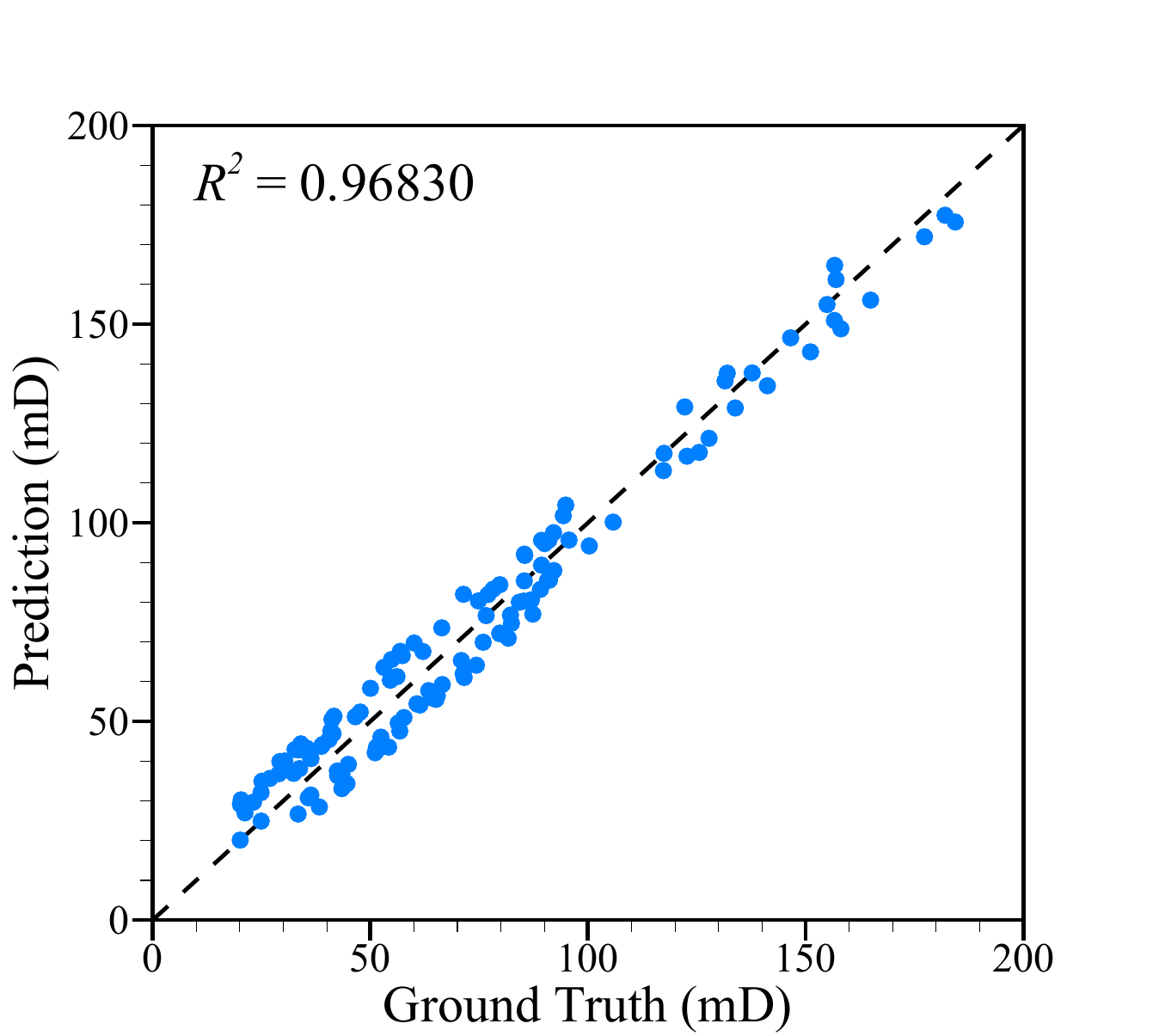}
            \put(5,90){\bfseries a}
        \end{overpic}
    \end{subfigure}
    \hfill
    \begin{subfigure}{0.33\textwidth}
        \centering
        \begin{overpic}[width=\textwidth]{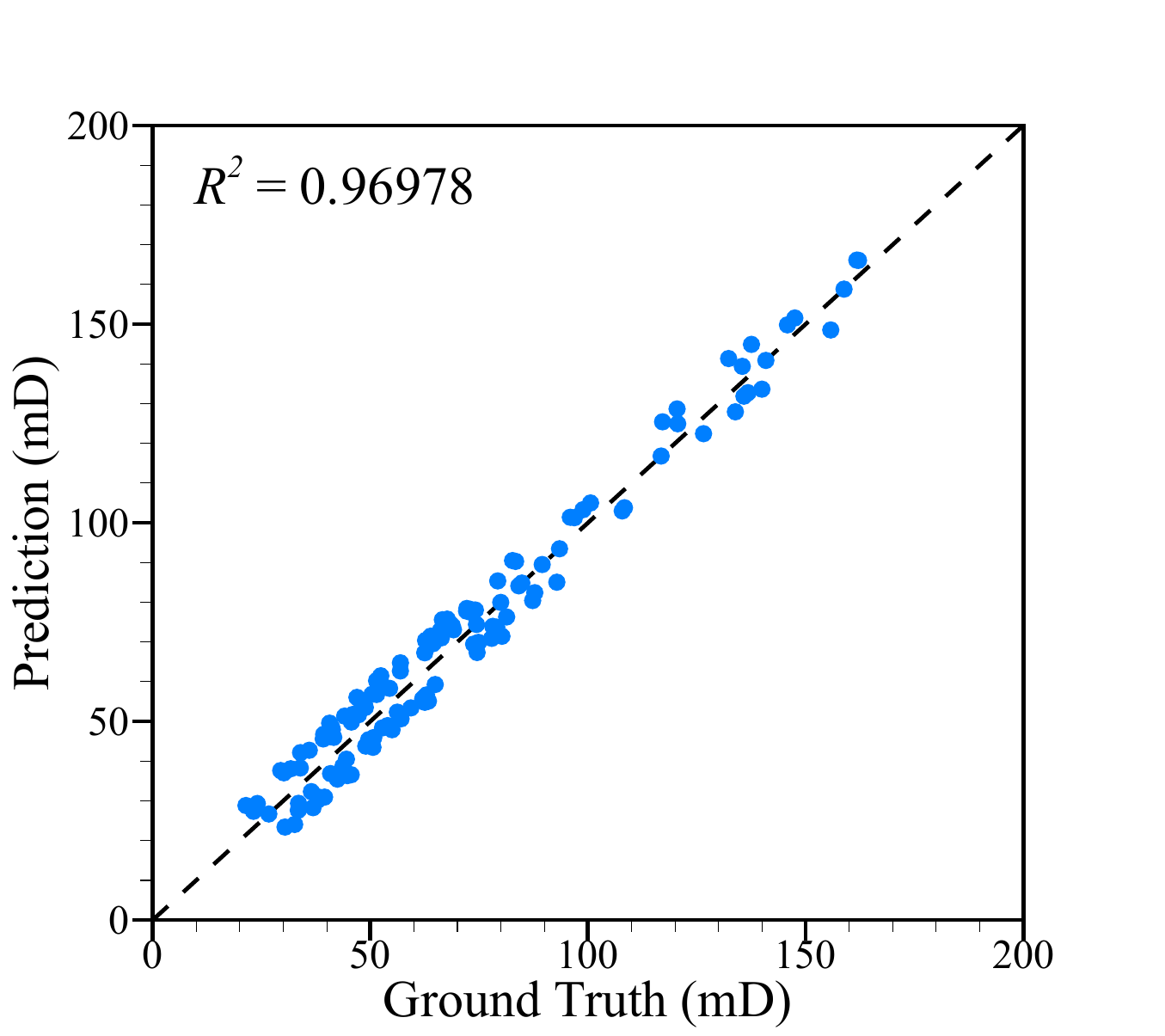}
            \put(5,90){\bfseries b}
        \end{overpic}
    \end{subfigure}
    \hfill
    \begin{subfigure}{0.33\textwidth}
        \centering
        \begin{overpic}[width=\textwidth]{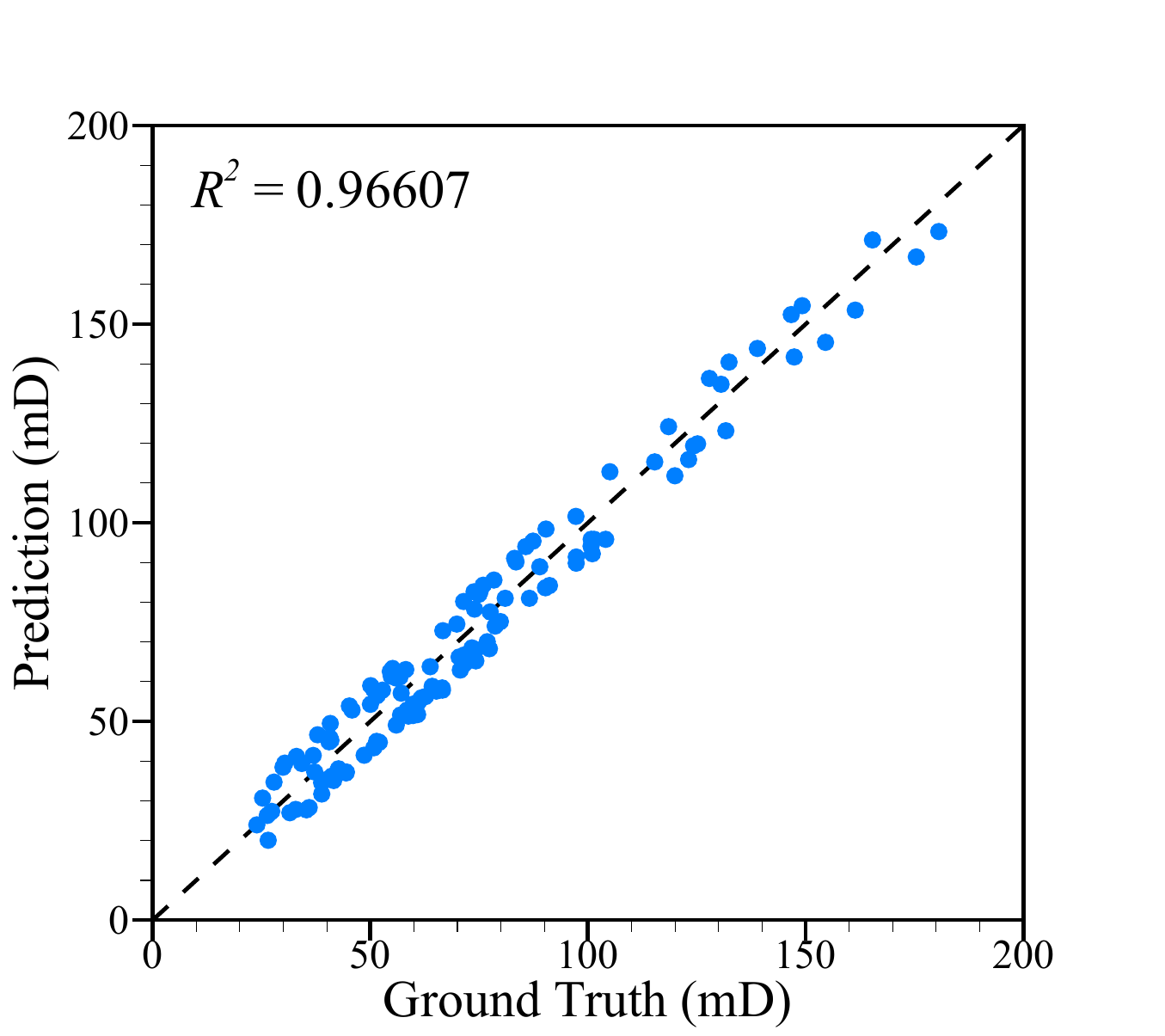}
            \put(5,90){\bfseries c}
        \end{overpic}
    \end{subfigure}
    \caption{$R^2$ plots for the test set (375 data) using the proposed approach for the classification of multi-sized images. The results are individually shown for \textbf{a} images of size $40^3$ (125 data), \textbf{b} images of size $48^3$ (125 data), and \textbf{c} images of size $56^3$ (125 data)}
    \label{Fig51}
\end{figure*}

As illustrated in Fig. \ref{Fig41}, the success of our approach is evident in the $R^2$ score, 0.96809, obtained for the test set (e.g., 375 data). Additionally, Fig. \ref{Fig51} specifically showcases the $R^2$ scores for the test set but individualized for each cubic size (i.e., $40^3$, $48^3$, and $56^3$). As can be seen in Fig. \ref{Fig51}, the $R^2$ scores obtained are equal to 0.96830, 0.96978, and 0.96607, respectively for the cubic digital porous media of sizes $40^3$, $48^3$, and $56^3$. The range of $R^2$ scores for the three different sizes remains at an excellent level, demonstrating that our FNO-based framework is robust and not overfitted to any specific size. Regarding the speedup achieved using the proposed deep learning framework, it predicts the permeability of the test set approximately in 18 seconds using our GPU machine. In contrast, computing the permeability of the same data with our in-house Lattice Boltzmann Method code, developed in C++ programming language, needs approximately 27 minutes on a single Intel(R) Core processor with a clock rate of 2.30 GHz. As a result, the average speedup factor we have accomplished is approximately 90 times faster compared to our conventional numerical solver. It is important to mention that the reported speedup factor is highly dependent on the efficiency of the numerical solver and the computing power. For instance, our numerical solver is a custom that operates on a single central processing unit. Modern and commercial applications (e.g., COMSOL and GeoDict) are significantly faster compared to our C++ code.


\begin{table*}
\caption{$R^2$ score of the test set for different mode numbers of the proposed FNO-based framework}
\begin{tabular}{llllllllll}
\hline
\hline
Number of modes in each dimension & 2 & 3 & 4 & 5 & 6 & 7 & 8 & 9 & 10 \\ \hline
$R^2$ score & 0.96809 & 0.15416 & 0.26757 & 0.26361 & 0.23325 & 0.38789 & 0.40433 & 0.31773 & 0.28839 \\
\hline
\hline
\end{tabular}
\label{Table1}
\end{table*}

\begin{figure*}
	\centering 
    \begin{subfigure}{0.33\textwidth}
        \centering
        \begin{overpic}[width=\textwidth]{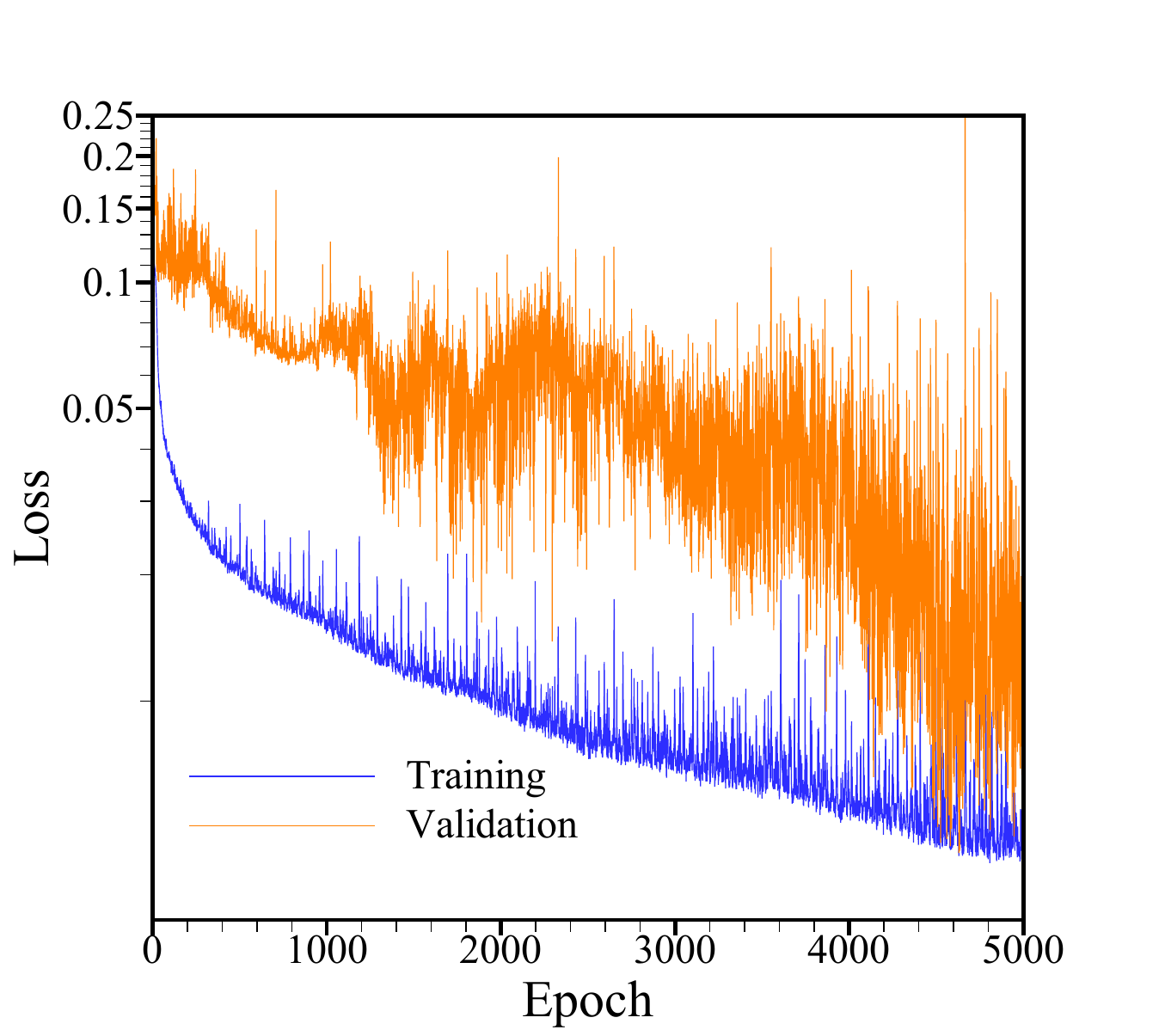}
            \put(5,90){\bfseries a}
        \end{overpic}
    \end{subfigure}
    \hfill
    \begin{subfigure}{0.33\textwidth}
        \centering
        \begin{overpic}[width=\textwidth]{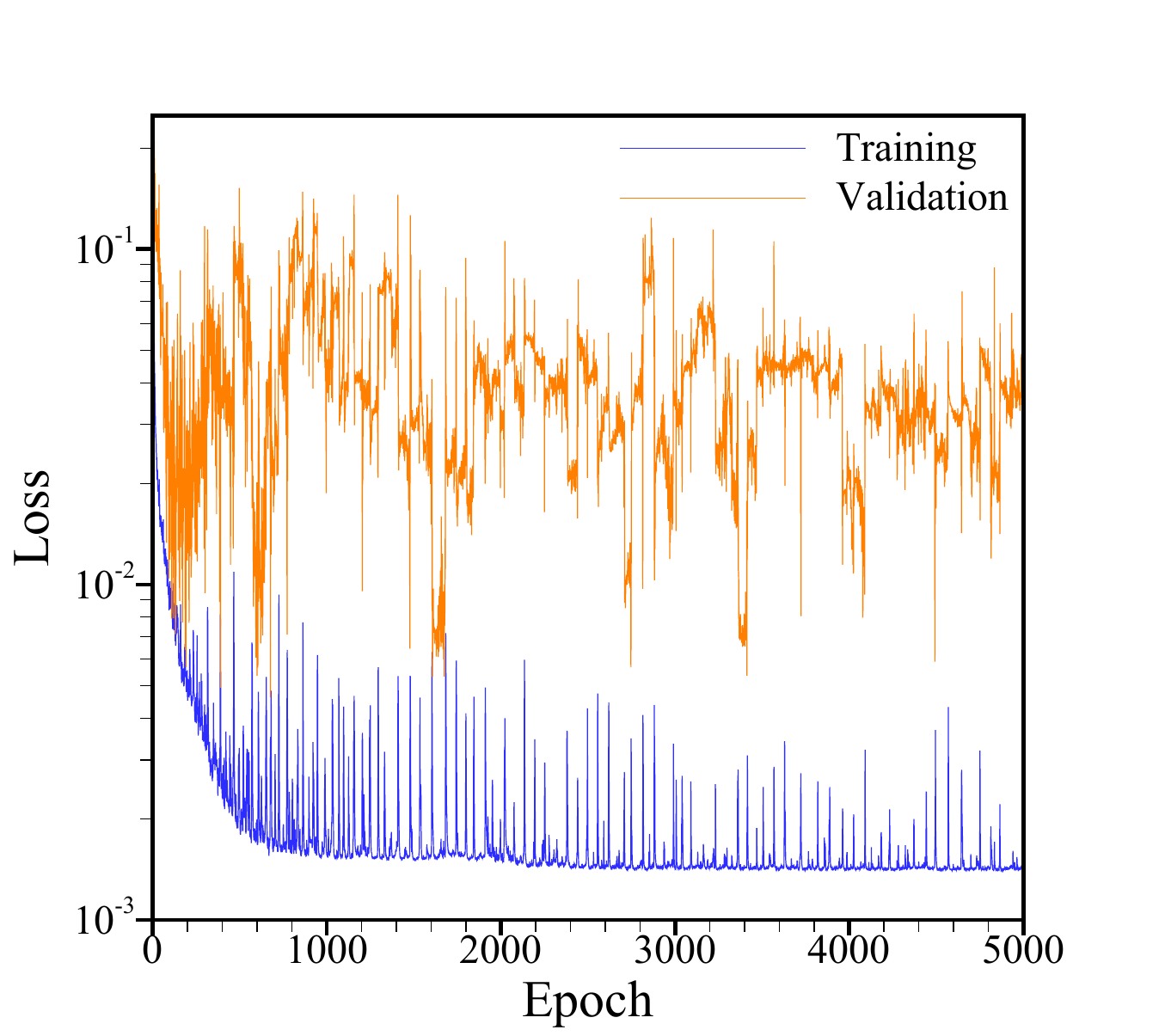}
            \put(5,90){\bfseries b}
        \end{overpic}
    \end{subfigure}
    \hfill
    \begin{subfigure}{0.33\textwidth}
        \centering
        \begin{overpic}[width=\textwidth]{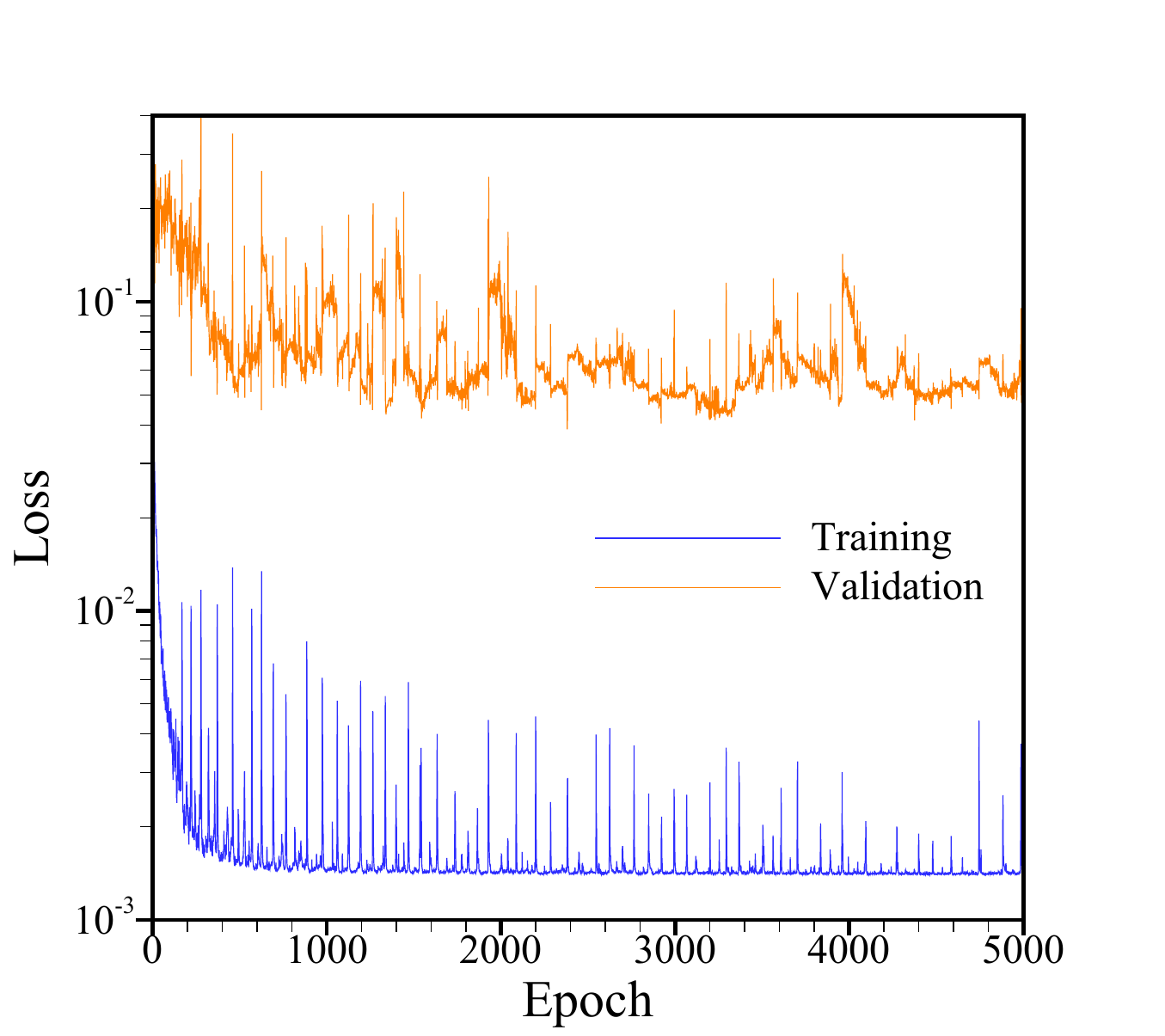}
            \put(5,90){\bfseries c}
        \end{overpic}
    \end{subfigure}
	\caption{Evolution of the loss function for the validation and training sets for the choice of
    \textbf{a} $m_{\text{max},1}=m_{\text{max},2}=m_{\text{max},3}=2$, \textbf{b} $m_{\text{max},1}=m_{\text{max},2}=m_{\text{max},3}=7$, and \textbf{c} $m_{\text{max},1}=m_{\text{max},2}=m_{\text{max},3}=10$}
	\label{Fig7}
\end{figure*}


\begin{figure*}
    \centering
    \begin{subfigure}{0.33\textwidth}
        \centering
        \begin{overpic}[width=\textwidth]{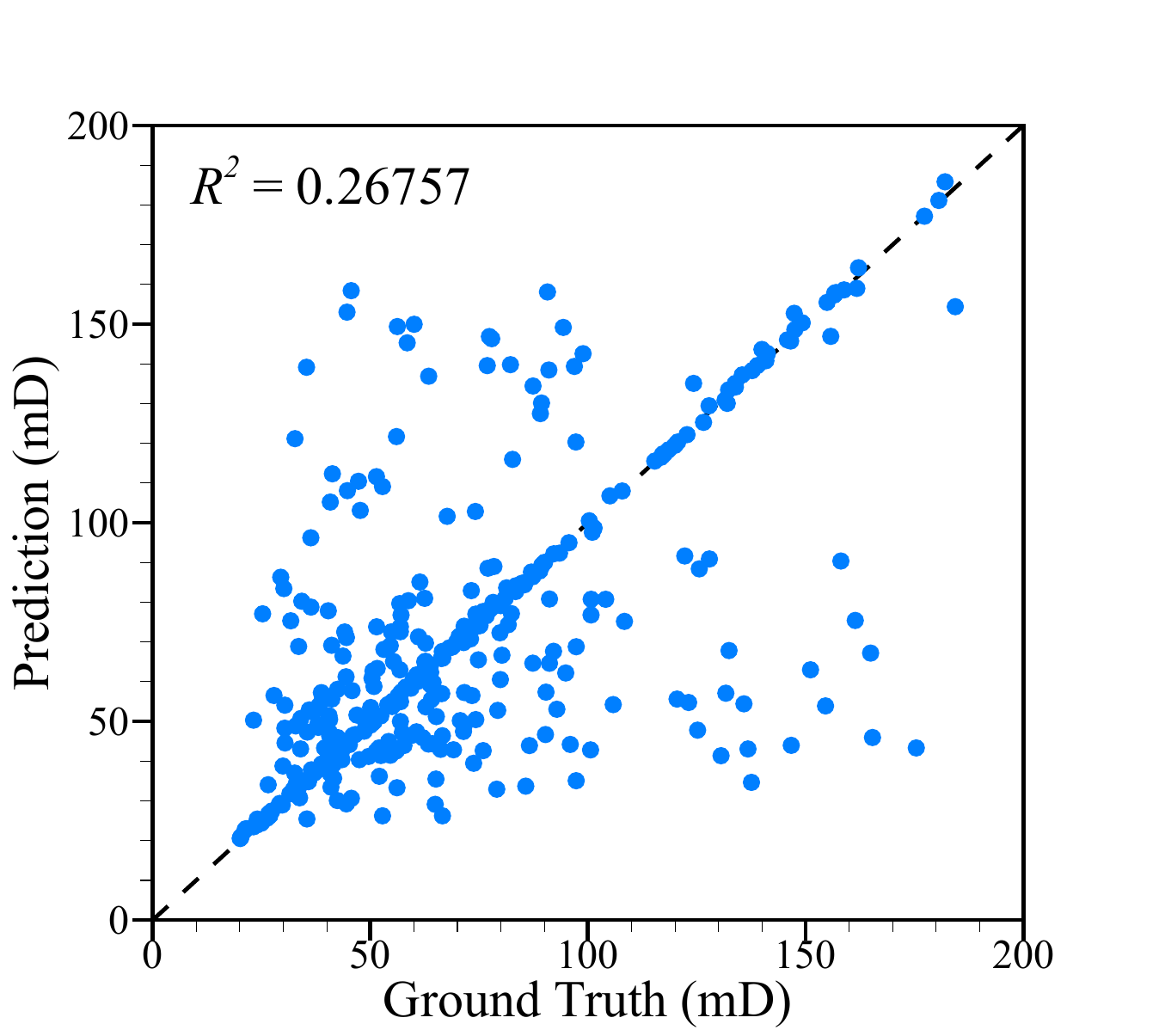}
            \put(5,90){\bfseries a}
        \end{overpic}
    \end{subfigure}
    \hfill
    \begin{subfigure}{0.33\textwidth}
        \centering
        \begin{overpic}[width=\textwidth]{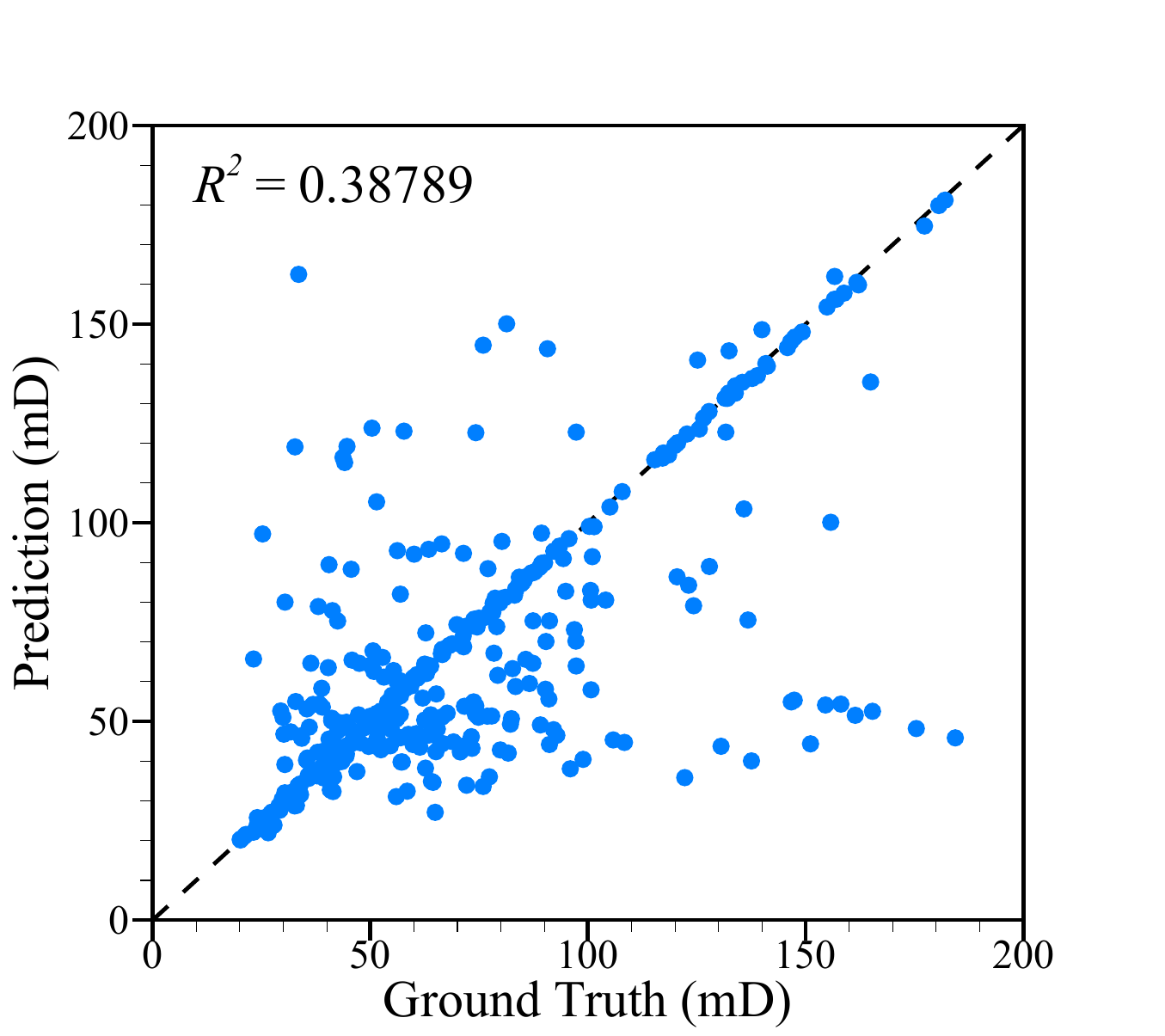}
            \put(5,90){\bfseries b}
        \end{overpic}
    \end{subfigure}
    \hfill
    \begin{subfigure}{0.33\textwidth}
        \centering
        \begin{overpic}[width=\textwidth]{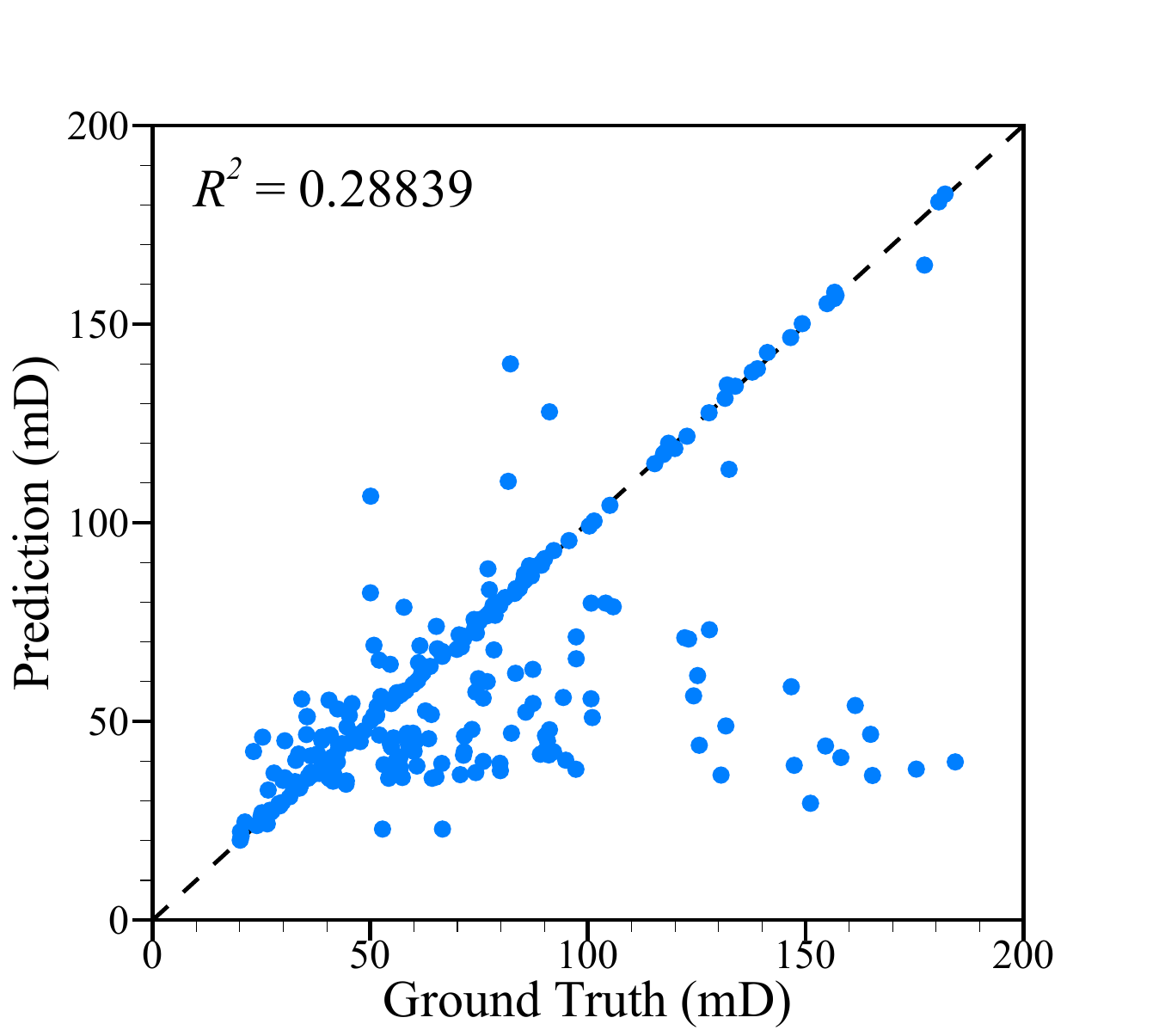}
            \put(5,90){\bfseries c}
        \end{overpic}
    \end{subfigure}
    \caption{$R^2$ plots for the test set (375 data) using the proposed approach for the classification of multi-sized images for the choice of \textbf{a} $m_{\text{max},1}=m_{\text{max},2}=m_{\text{max},3}=4$, \textbf{b} $m_{\text{max},1}=m_{\text{max},2}=m_{\text{max},3}=7$, and \textbf{c} $m_{\text{max},1}=m_{\text{max},2}=m_{\text{max},3}=10$}
    \label{Fig600}
\end{figure*}


\begin{figure*}
    \centering
    \begin{subfigure}{0.45\textwidth}
        \centering
        \begin{overpic}[width=\textwidth]{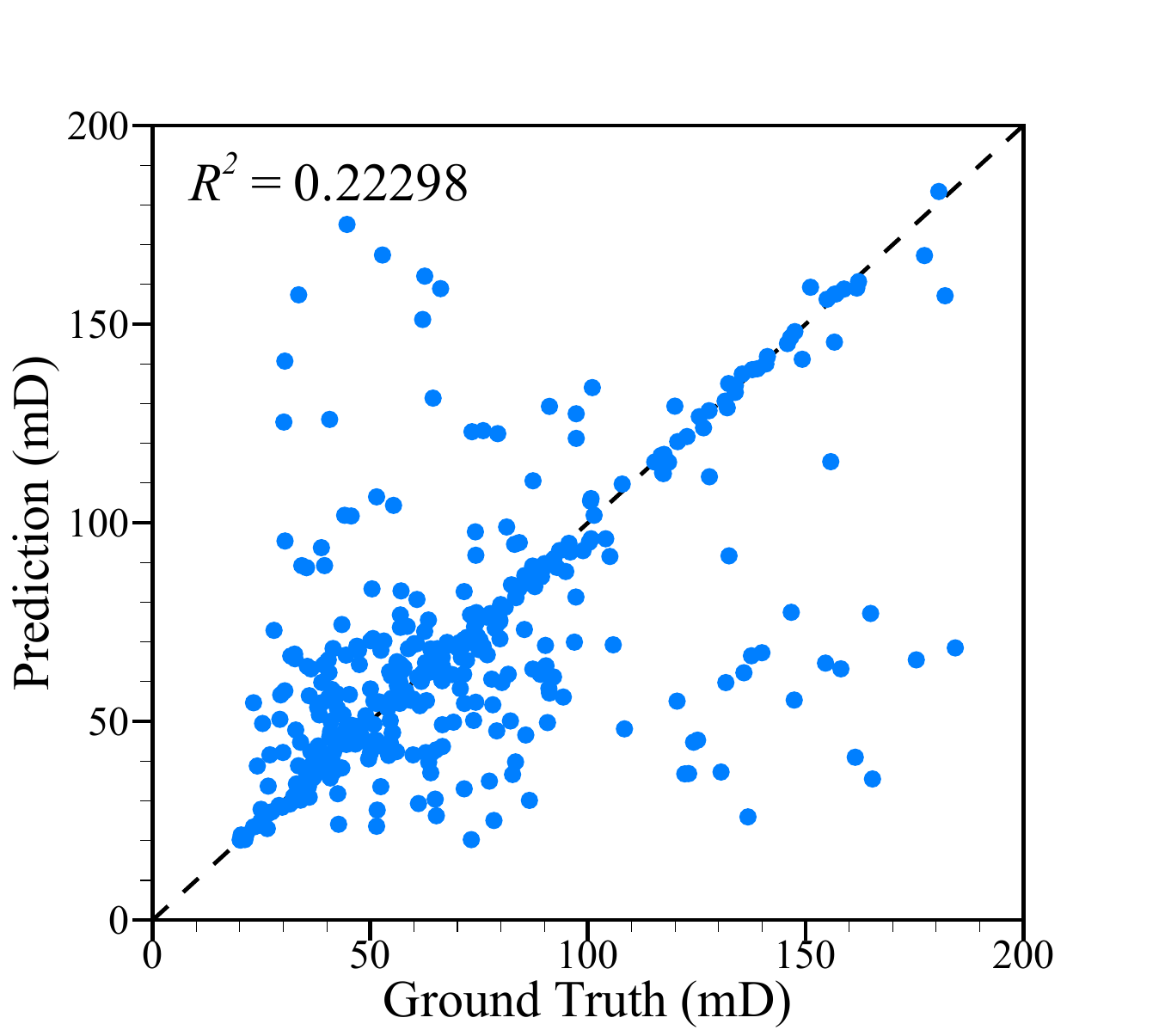}
            \put(5,90){\bfseries a}
        \end{overpic}
    \end{subfigure}
    \hfill
    \begin{subfigure}{0.45\textwidth}
        \centering
        \begin{overpic}[width=\textwidth]{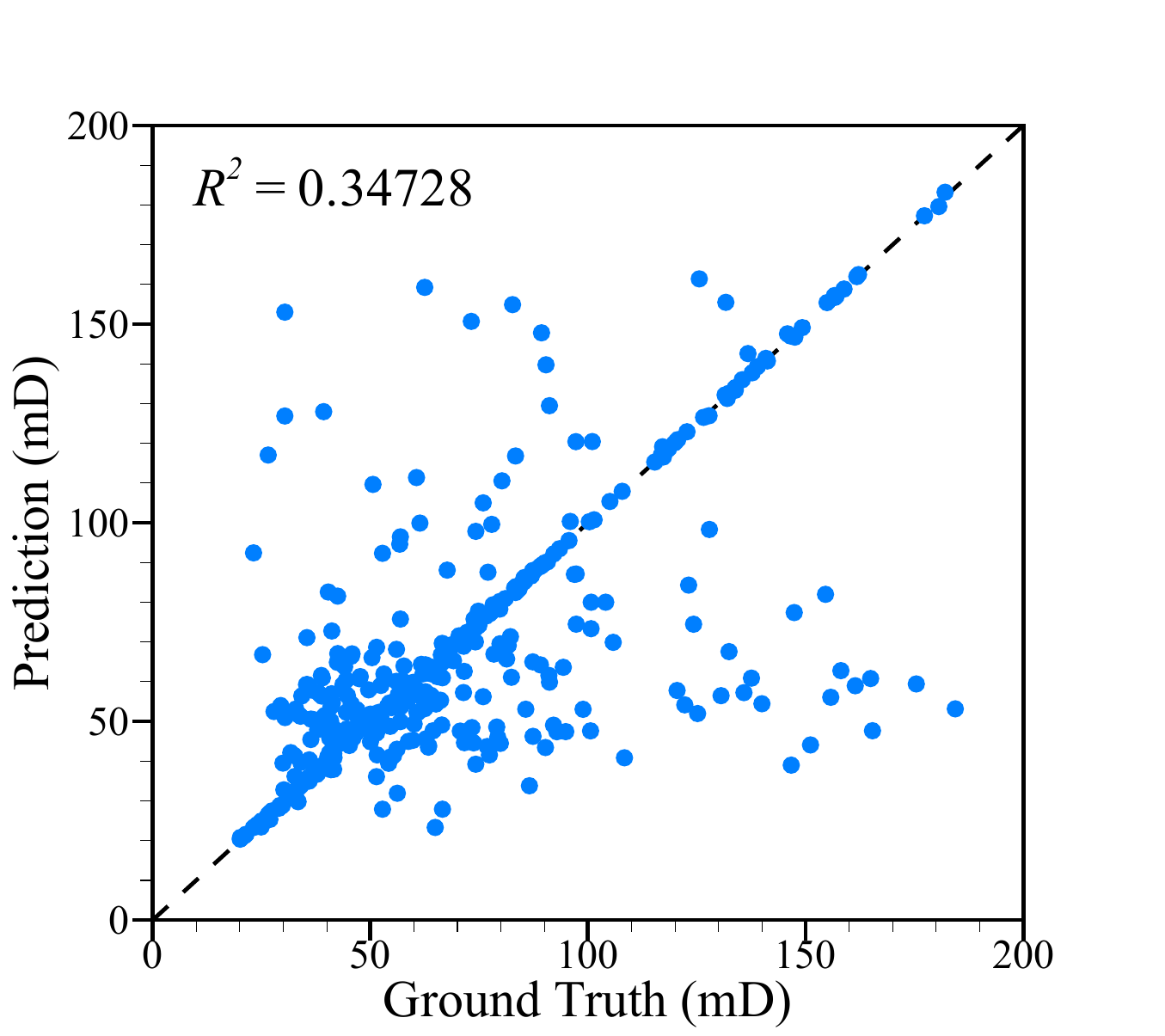}
            \put(5,90){\bfseries b}
        \end{overpic}
    \end{subfigure}
    \caption{$R^2$ plots for the test set (375 data) using the proposed approach for the classification of multi-sized images for the choice of \textbf{a} $m_{\text{max},1}=m_{\text{max},2}=2$ and $m_{\text{max},3}=10$, and the choice of \textbf{b} $m_{\text{max},1}=2$, and $m_{\text{max},2}=m_{\text{max},3}=10$}
    \label{Fig601}
\end{figure*}


\subsection{Number of Fourier modes in each dimension}
\label{Sect522}

Our deep learning experiments demonstrate that there is a critical interplay between the number of modes (i.e., $m_{\text{max},1}$, $m_{\text{max},2}$, and $m_{\text{max},3}$) set in the proposed FNO framework and the tendency for overfitting during the training procedure. Accordingly, setting the number of modes beyond 2 leads to a severe divergence between the training and validation loss. This fact can be observed in Fig. \ref{Fig7} when we set $m_{\text{max},1}=7$, $m_{\text{max},2}=7$, and $m_{\text{max},3}=7$ or $m_{\text{max},1}=10$, $m_{\text{max},2}=10$, and $m_{\text{max},3}=10$. The reported results indicate that the number of modes plays a critical role in the FNO model generalization. A further survey of the influence of the number of modes in the FNO configuration is performed by varying the number of modes in all three principal directions, from 2 to 10, and the obtained $R^2$ scores are tabulated in Table \ref{Table1}. Accordingly, the optimal mode configuration for avoiding overfitting is 2, as the divergence between the validation and training loss is minimized. Consequently, a careful selection of the number of modes in the FNO units is necessary to make the deep learning framework robust and reliable for the image classification application. The consequence of this scenario is observable in Fig. \ref{Fig600}, where we plot the $R^2$ score for the test sets, for example, for the choice of $m_{\text{max},1} = m_{\text{max},2} = m_{\text{max},3}=4$, $m_{\text{max},1}= m_{\text{max},2}= m_{\text{max},3}=7$, and $m_{\text{max},1} = m_{\text{max},2}=m_{\text{max},3}=10$. In all of these cases, the $R^2$ scores obtained for the prediction of the permeability of the porous media in the test set are less than 0.4.

We perform two other experiments. In the first one, we set only one mode (e.g., $m_{\text{max},3}$) to 10 ($m_{\text{max},3}=10$) and the other two modes to 2 (i.e., $m_{\text{max},1}=2$ and $m_{\text{max},2}=2$). In the second one, we set only two modes (e.g., $m_{\text{max},2}$ and $m_{\text{max},3}$) to 10 and the reminder mode to 2 (i.e., $m_{\text{max},1}=2$). The outputs of these two experiments are illustrated in Fig. \ref{Fig601}. As can be seen in Fig. \ref{Fig601}, the resulting $R^2$ scores of the test set are equal to 0.22298 and 0.34728, respectively, for the first and second experiments. Accordingly, we conclude that even increasing one mode beyond 2 drastically negatively affects the performance of the proposed FNO framework for the current application. Hence, the main challenge of working with the proposed network is its high sensitivity to the number of modes. As discussed in this subsection, changing even the number of modes in one dimension leads to overfitting of the network on the training data and a lack of efficiency in predicting the test data, consequently resulting in a lack of generalizability.


\subsection{Channel width of FNOs}
\label{Sect523}
We further analyze the impact of different channel widths on the performance of the introduced deep learning framework. Based on our machine learning experiments, $R^2$ scores obtained for the channel width of 8, 27, 64, and 125 are 0.49904, 0.81618, 0.96815, and 0.94457, respectively. When the channel width decreases from 64 to 27 or to 8, a significant drop in the $R^2$ score is observed. Notably, increasing the channel width beyond 64 to 125 also leads to a slight decrease in the precision of permeability predictions.

As discussed in Sect. \ref{Sect23}, the choice of channel width is directly related to the number of trainable parameters, which are 30897, 163783, 828673, and 3096531 for each respective channel width. Moreover, the channel width also determines the size of the max pooling, representing the size of the global feature vector. Hence, optimizing channel width is critical. Small channel width leads to poor performance, whereas large channel width imposes high computational costs and memory allocation without necessarily a significant performance improvement.

\subsection{Number of FNO units}
\label{Sect524}
We investigate the effect of varying the number of FNO units (see Fig. \ref{Fig1}). Deep learning experiments are conducted using one, two, three, four, and five units to assess the impact on the introduced FNO performance. By computing the $R^2$ score across the test set, we realize that there is no significant improvement in the prediction accuracy. $R^2$ score for the FNO configuration with one, two, three, four, and five units are respectively, 0.82767, 0.91703, 0.96813, 0.96759, and 0.97818. Hence, adding more units (beyond 3) and making the network deeper does not have a remarkable effect on the prediction accuracy. However, the number of trainable parameters and consequently, the computational cost and required GPU memory (e.g., RAM) escalated by adding FNO units. For example, 820353, 824513, 828673, 832833, and 836993 are, respectively, the number of trainable parameters of the model with one, two, three, four, and five layers of FNOs.

\subsection{Activation functions}
\label{Sect525}
We give particular attention to the effect of choosing an activation function on the prediction ability of our FNO model. In the primary setup, we configure all layers to employ the ReLU activation function, except the last layer of the classifier, where we utilize a sigmoid function. We implement two alternative setups. In the first one, we alter the activation function of the last layer to ReLU, this configuration results in a drastic reduction in the $R^2$ score of the test set, regardless of if the output permeability is normalized between 0 and 1. In the second setup, we replace the activation function in all layers with sigmoid. As a consequence of this setup, a slight decrease in performance is indicated, as $R^2$ score of 0.91478 is obtained for the test set. Note that the training procedure becomes slower in this setup, as the derivative of the sigmoid function results in a more complicated computation graph compared to that one output by the derivation of the ReLU function.

\subsection{Static max pooling versus static average pooling}
\label{Sect526}
Within the context of capturing global features in the proposed FNO-based framework, we explore the efficacy of implementing static average pooling as an alternative to static max pooling. Our machine learning experiment yields a $R^2$ score of 0.94478 in this case, demonstrating a marginal diminishment in the network performance compared to the presence of static max pooling. As supported by the literature \citep{qi2017pointnet,NIPS2017PointNetPlus,CFDPointNet,KASHEFIPIPNJCP,KashefiPIPNElasticity}, max pooling is a preferred technique for classification tasks compared to average pooling. Our finding shows a similar pattern for the introduced FNO-based framework.


\begin{figure*}
    \centering
    \begin{subfigure}{0.24\textwidth}
        \centering
        \begin{overpic}[width=\textwidth]{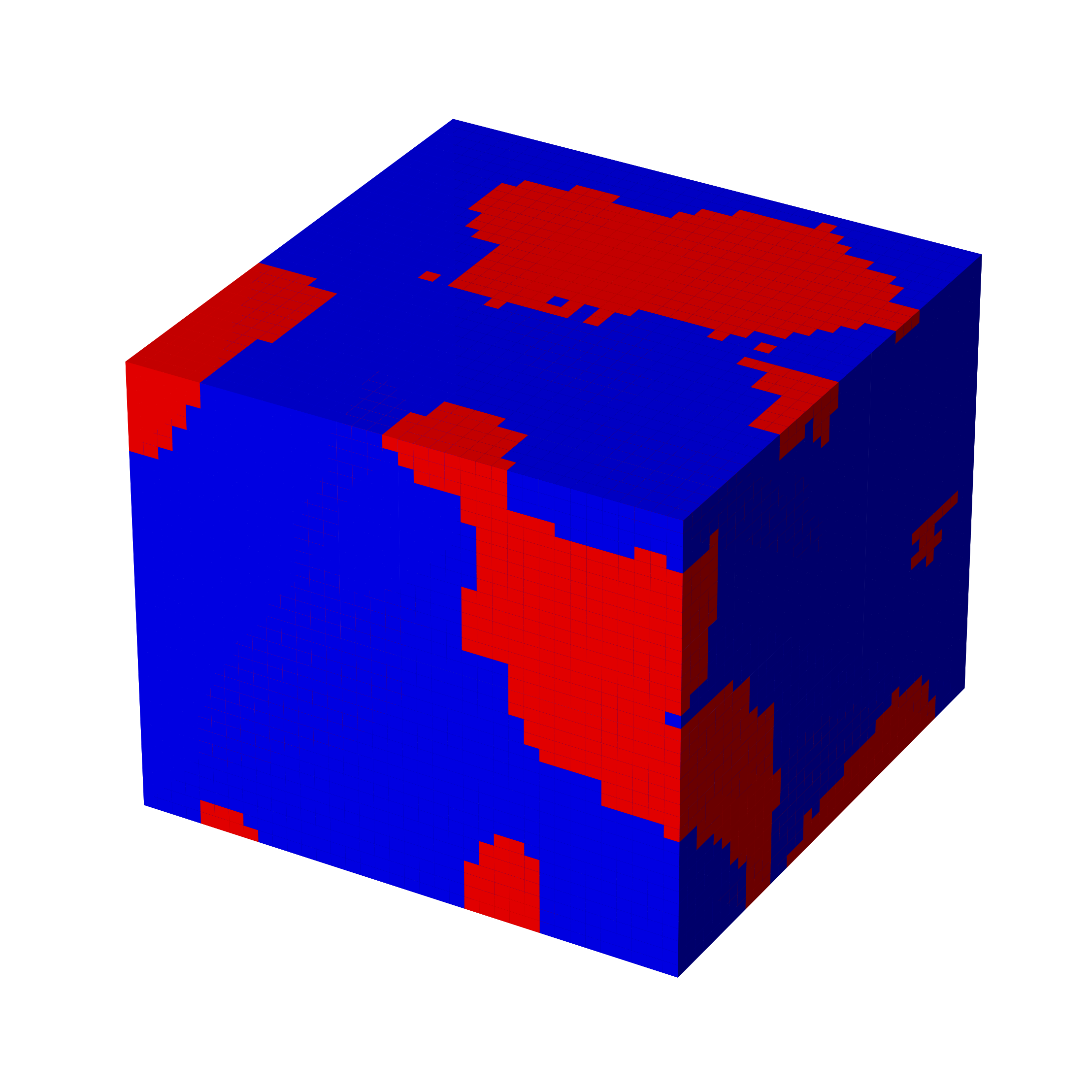}
            \put(5,90){\bfseries a}
        \end{overpic}
    \end{subfigure}
    \begin{subfigure}{0.24\textwidth}
        \centering
        \begin{overpic}[width=\textwidth]{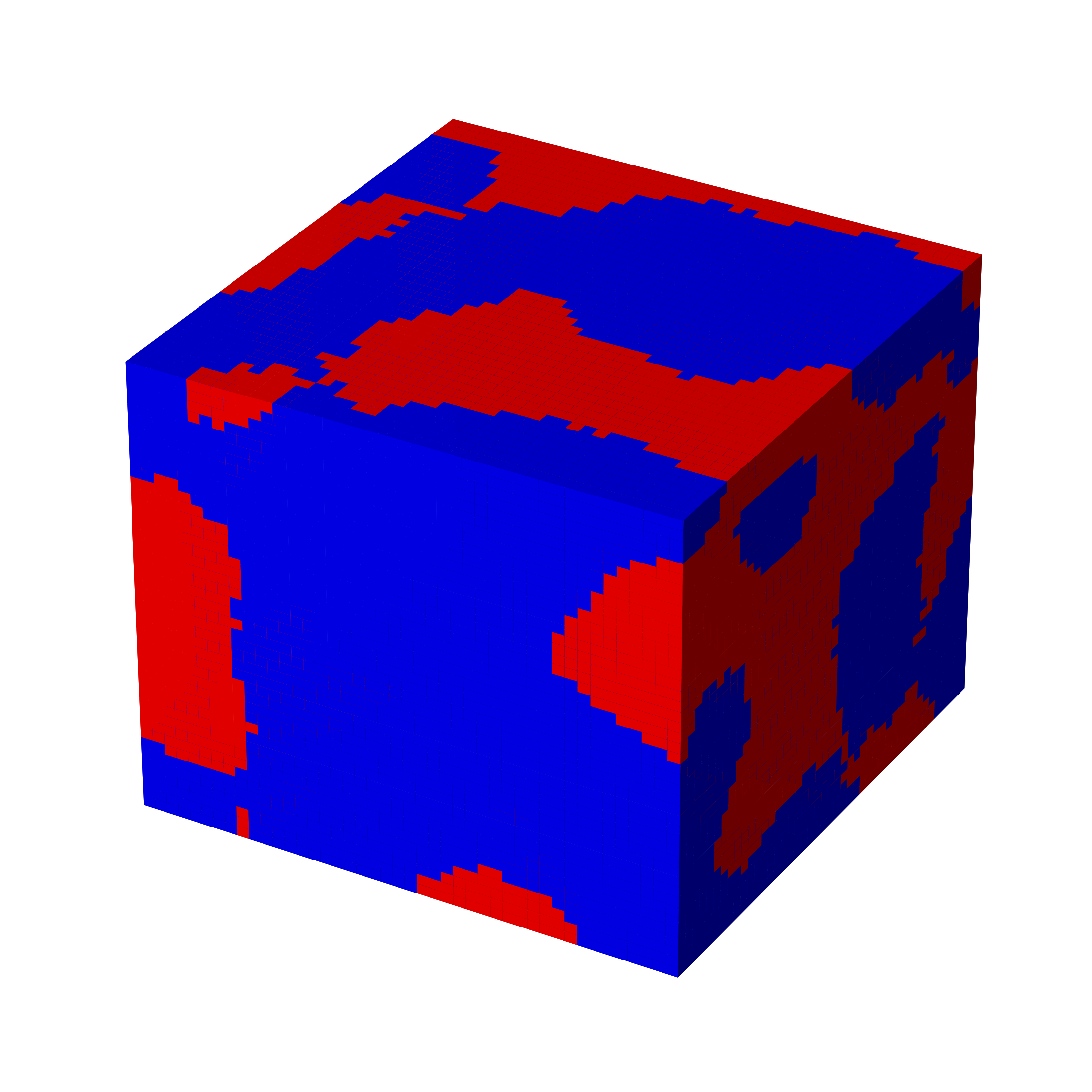}
            \put(5,90){\bfseries b}
        \end{overpic}
    \end{subfigure}
    \begin{subfigure}{0.24\textwidth}
        \centering
        \begin{overpic}[width=\textwidth]{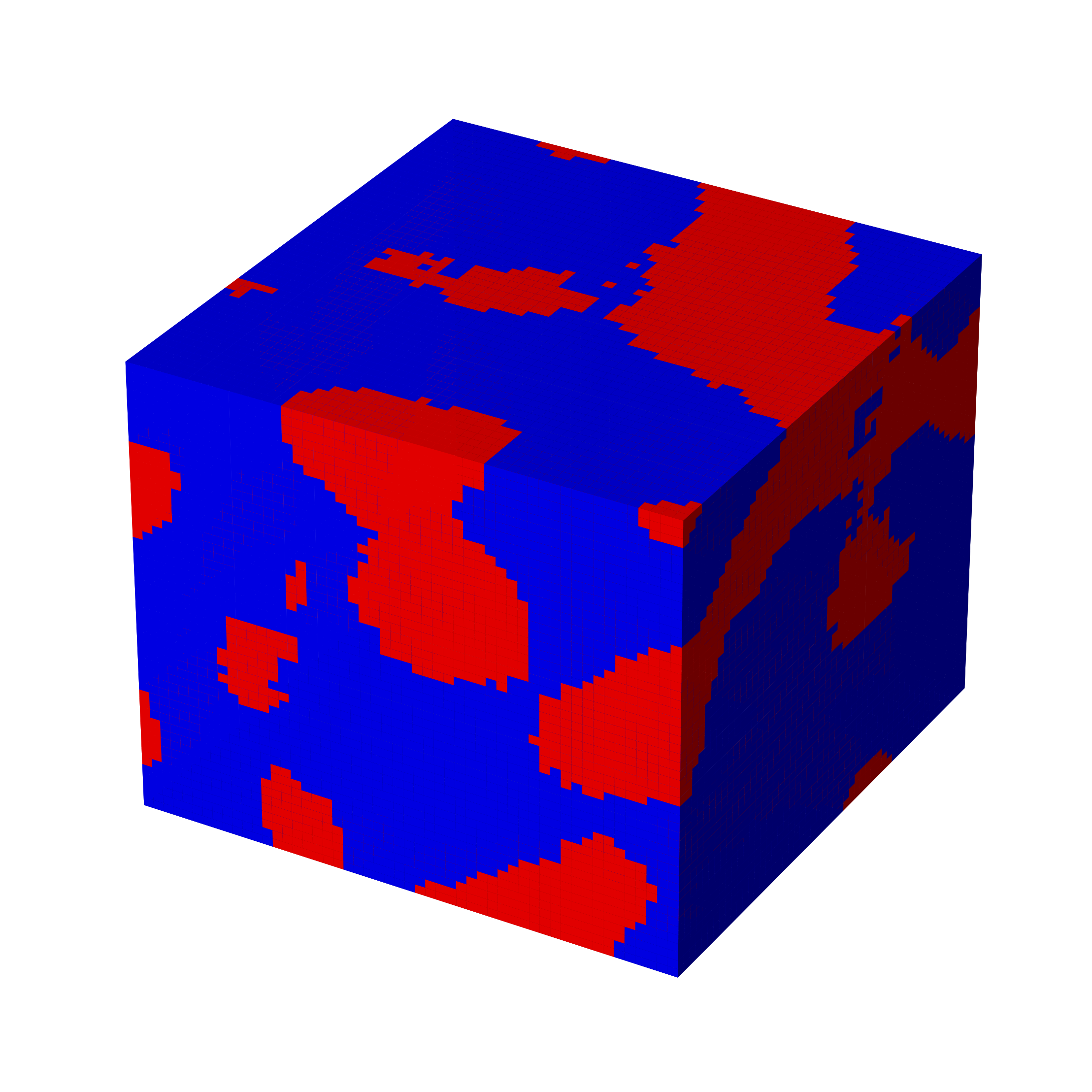}
            \put(5,90){\bfseries c}
        \end{overpic}
    \end{subfigure}
    \begin{subfigure}{0.24\textwidth}
        \centering
        \begin{overpic}[width=\textwidth]{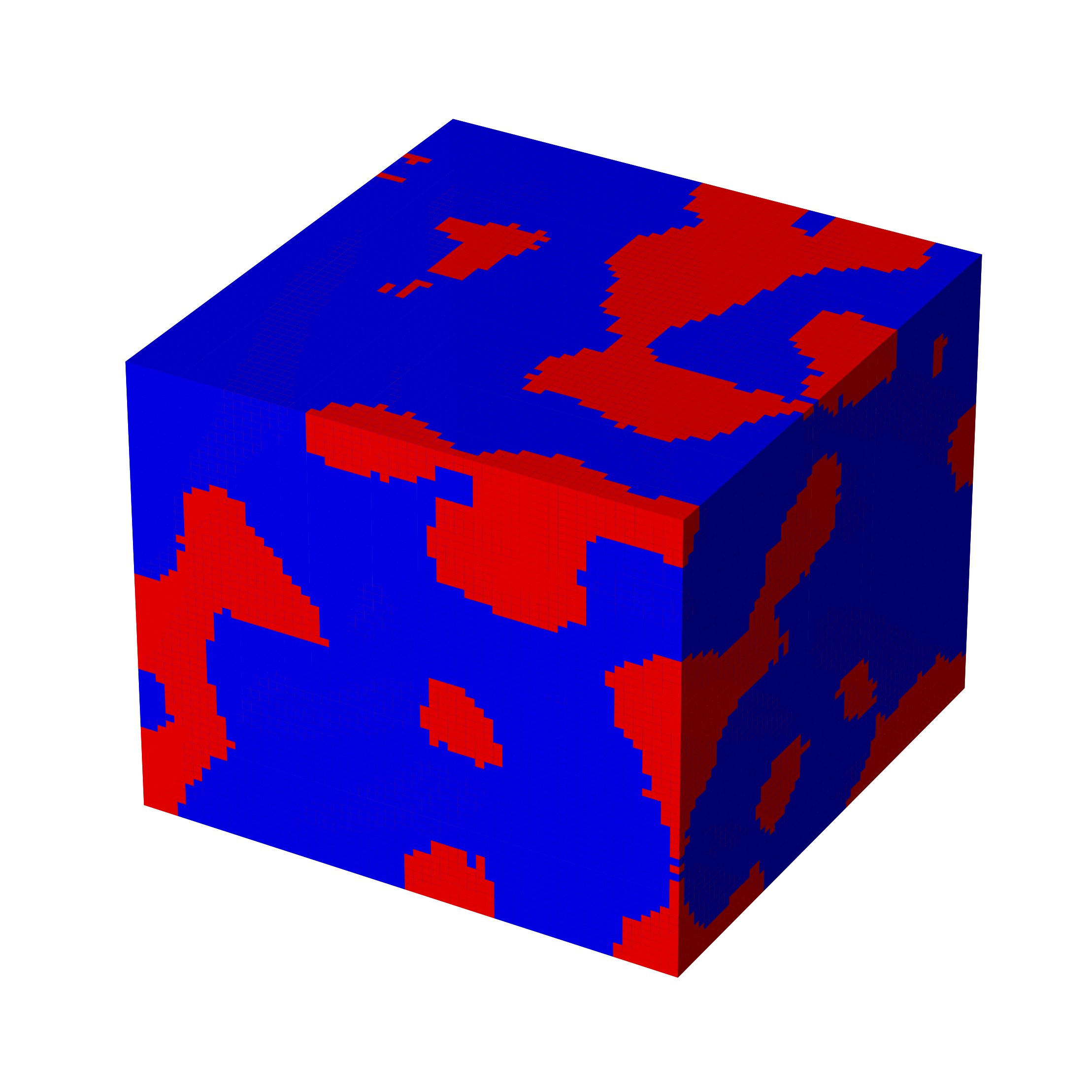}
            \put(5,90){\bfseries d}
        \end{overpic}
    \end{subfigure}
   \caption{A few examples of synthetically generated three-dimensional digital porous media for examining the generalizability of the proposed neural network; \textbf{a} an image of size $36^3$, \textbf{b} an image of size $44^3$, \textbf{c} an image of size $52^3$, and \textbf{d} an image of size $60^3$. Blue represents grain space, while red indicates pore space.}
    \label{Fig799}
\end{figure*}


\begin{figure*}
    \centering
    \begin{subfigure}{0.45\textwidth}
        \centering
        \begin{overpic}[width=\textwidth]{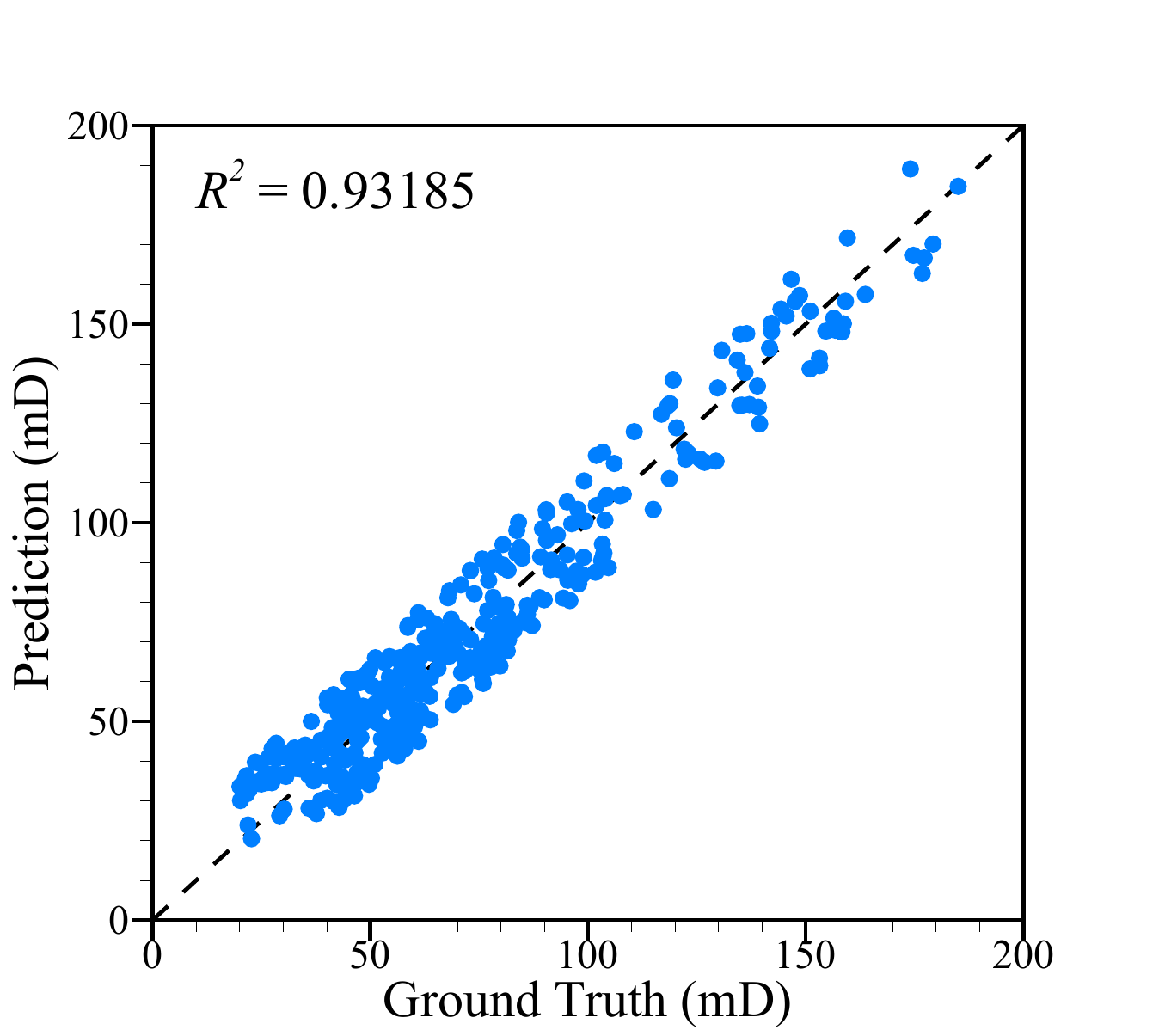}
            \put(5,90){\bfseries a}
        \end{overpic}
    \end{subfigure}
    \begin{subfigure}{0.45\textwidth}
        \centering
        \begin{overpic}[width=\textwidth]{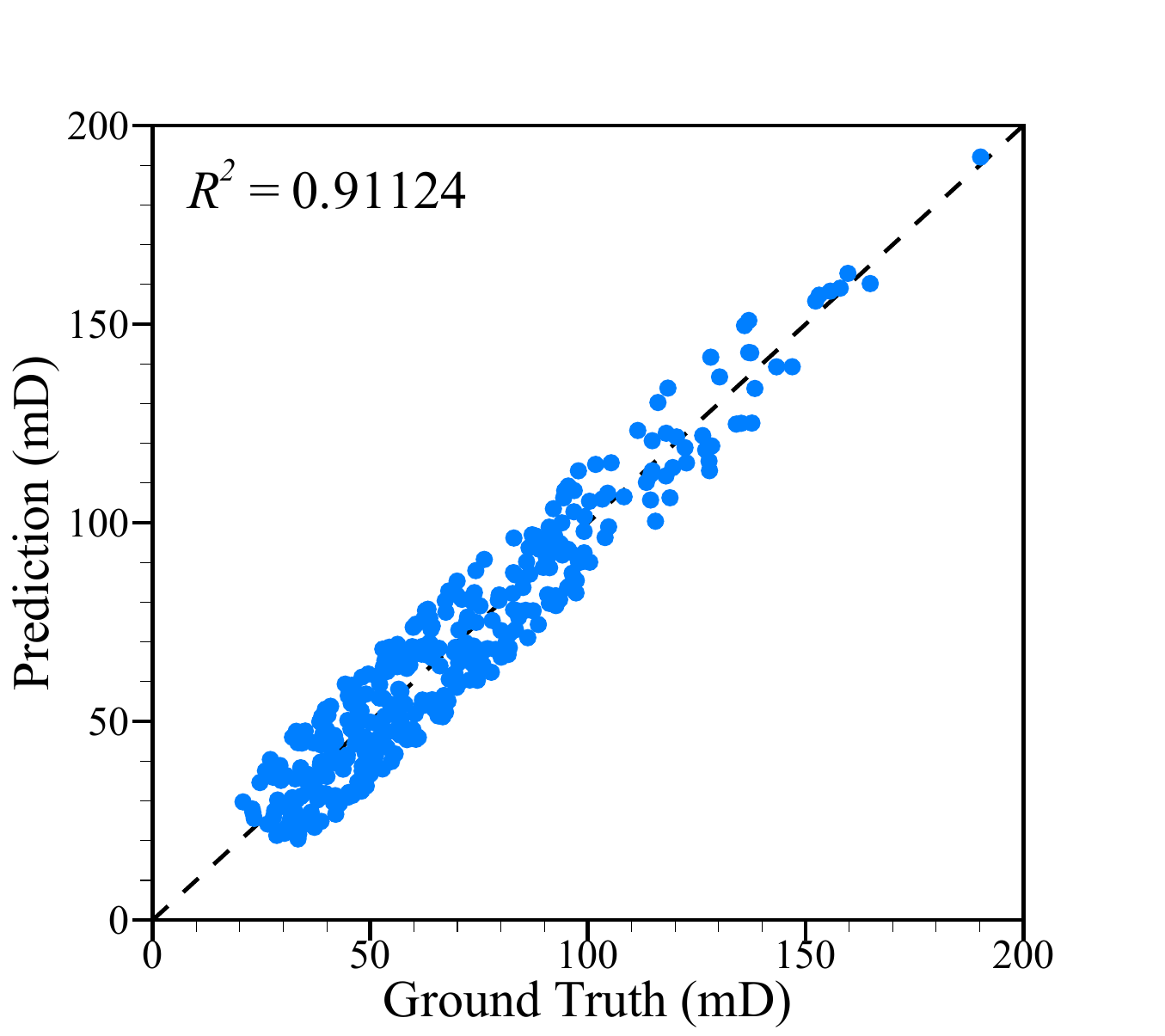}
            \put(5,90){\bfseries b}
        \end{overpic}
    \end{subfigure}
    \begin{subfigure}{0.45\textwidth}
        \centering
        \begin{overpic}[width=\textwidth]{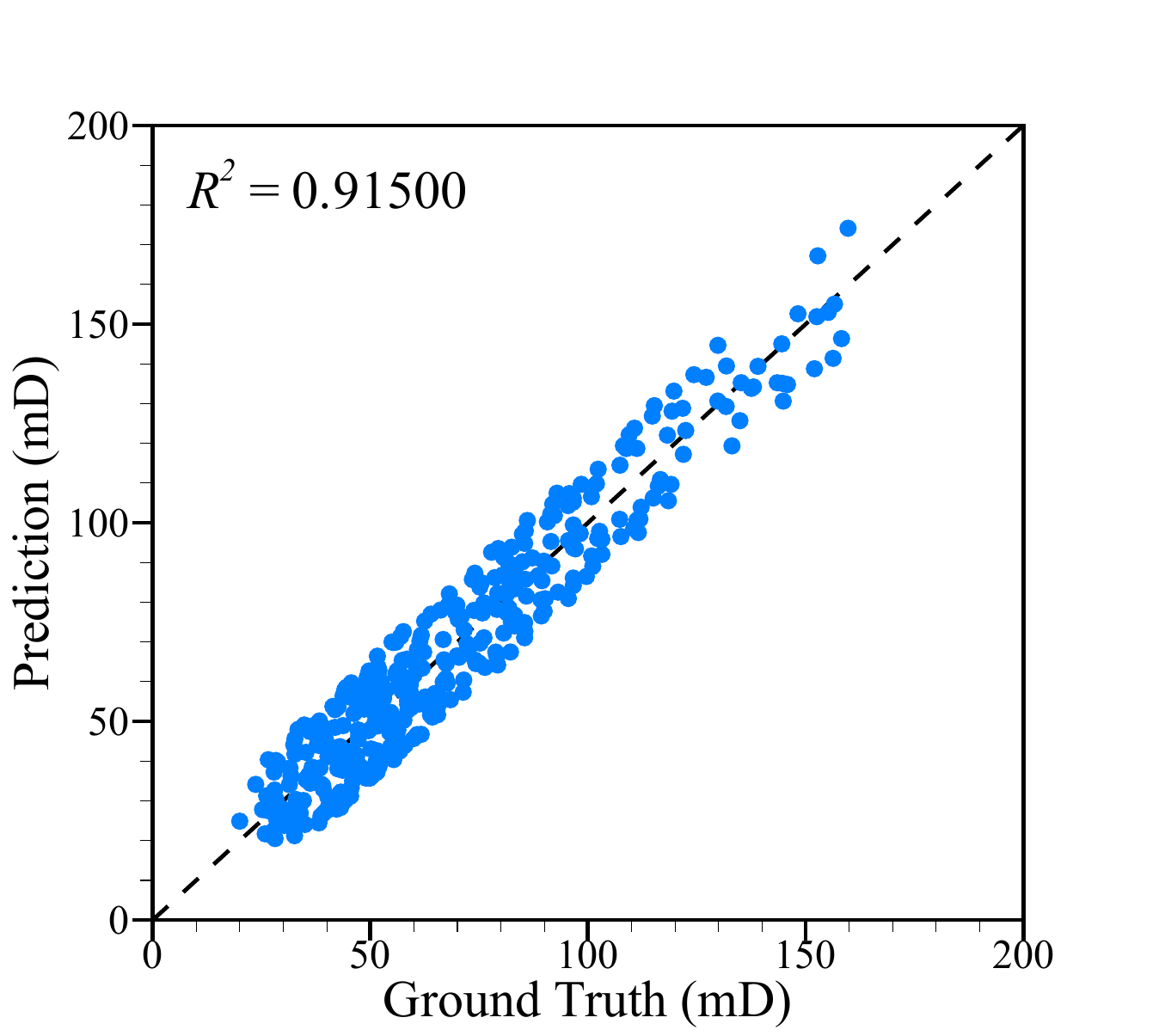}
            \put(5,90){\bfseries c}
        \end{overpic}
    \end{subfigure}
    \begin{subfigure}{0.45\textwidth}
        \centering
        \begin{overpic}[width=\textwidth]{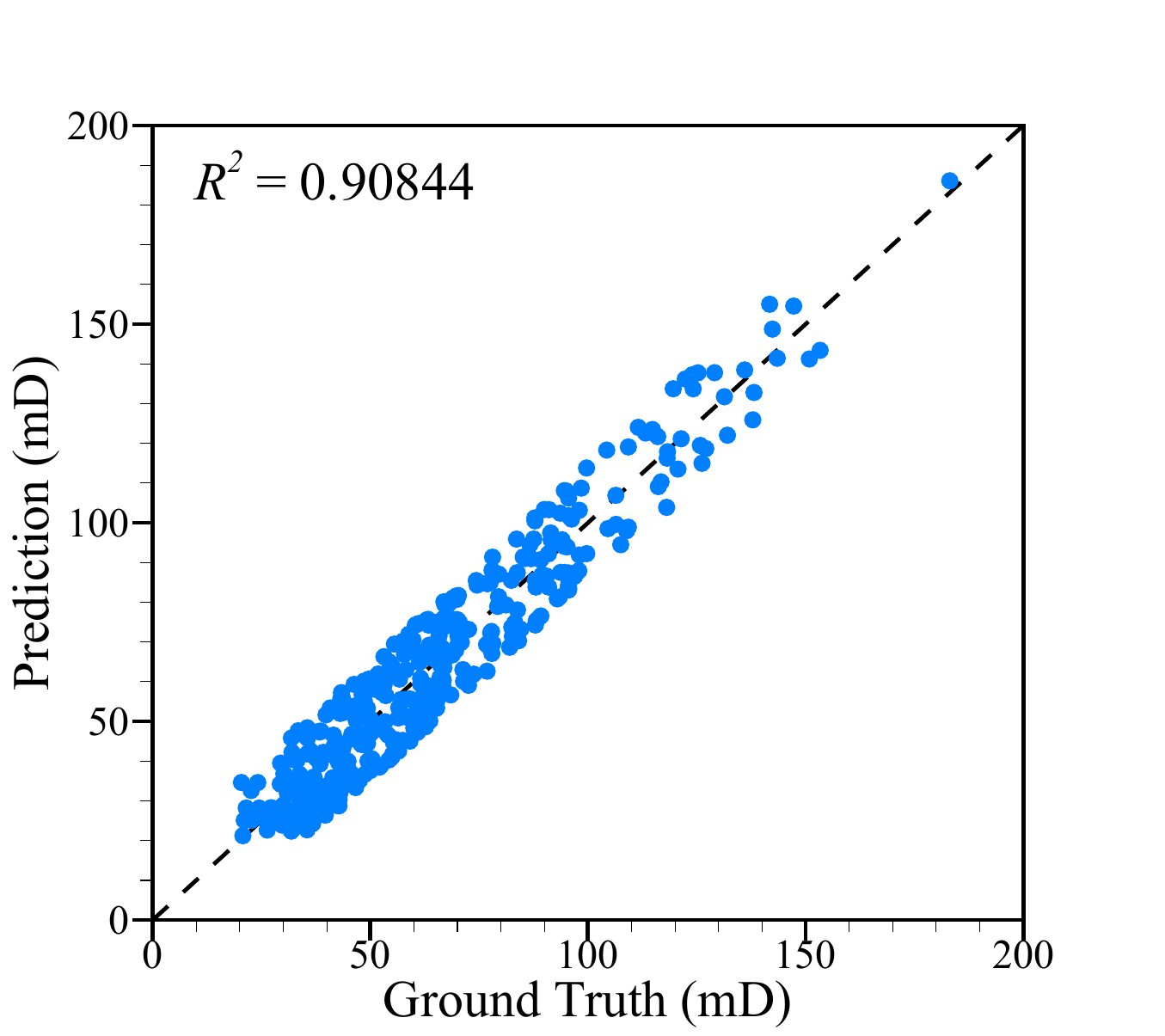}
            \put(5,90){\bfseries d}
        \end{overpic}
    \end{subfigure}
   \caption{$R^2$ plots demonstrating the generalizability of the proposed approach in classifying multi-sized images. The network, trained on images of sizes $40^3$, $48^3$, and $56^3$, is used to predict images of sizes \textbf{a} $36^3$ (375 data), \textbf{b} $44^3$ (375 data), \textbf{c} $52^3$ (375 data), and \textbf{d} $60^3$ (375 data).}
    \label{Fig801}
\end{figure*}

\subsection{Generalizability}
\label{Sect527}
In this subsection, we assess the generalizability ability of the proposed FNO-based framework. Note that the concept of generalizability in the context of the present work extends to the network's performance to predict the permeability of cubic porous media with unseen sizes. As discussed in Sect. \ref{Sect4}, the network was initially trained using porous media with cubic geometries of sizes $40^3$, $48^3$, and $56^3$. To examine the network capacity to generalize, we predict the permeability of porous media with sizes $36^3$, $44^3$, $52^3$, and $60^3$ with 375 cubes for each of these sizes using our pretrained FNO-based framework. Figure \ref{Fig799} shows a few examples of these synthetic data, generated for the purpose of examining the network generalizability. As shown in Fig. \ref{Fig801}, a slight decline is observed in the accuracy of permeability predictions for porous media with unseen sizes. However, the obtained \( R^{2} \) scores remain in an excellent range. These scores are 0.93185, 0.91124, 0.91500, and 0.90844 for the porous media sizes of \( 36^{3} \), \( 44^{3} \), \( 52^{3} \), and \( 60^{3} \), respectively. As another observation, the performance of our approach is marginally higher in predicting the permeability of unseen porous media with smaller cubic sizes. As highlighted in Fig. \ref{Fig801}, $R^2$ scores of porous media with sizes of $36^3$ are greater than ones with a size of $44^3$. A similar scenario occurs when we compare porous media of sizes $52^3$ and $60^3$. This can be attributed to the fact that, for smaller sizes, the fixed-size vector of the global feature encodes the features of smaller cubes more effectively. Moreover, note that the vector size is the same as the width channel. As a last comment in this subsection, to enhance the network's generalizability, a potential strategy could involve expanding the training dataset to include more than the initial three sets of geometry sizes.

\begin{figure*}
	\centering 
    \begin{subfigure}{0.47\textwidth}
        \centering
        \begin{overpic}[width=\textwidth]{Loss_mode2.pdf}
            \put(5,90){\bfseries a}
        \end{overpic}
    \end{subfigure}
    \hfill
    \begin{subfigure}{0.47\textwidth}
        \centering
        \begin{overpic}[width=\textwidth]{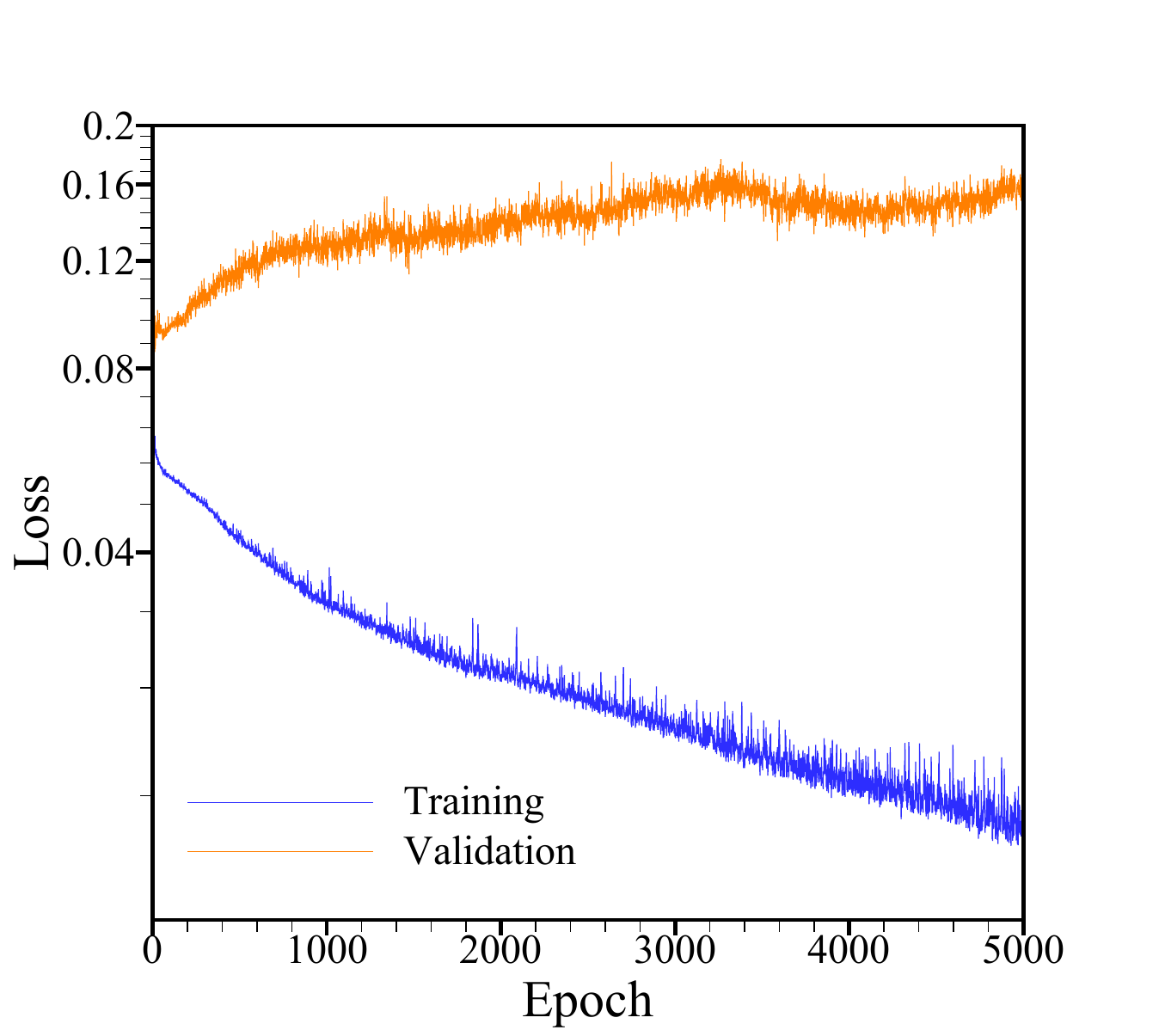}
            \put(5,90){\bfseries b}
        \end{overpic}
    \end{subfigure}
    \hfill
	\caption{Evolution of the loss function for the validation and training sets using \textbf{a} the proposed approach (see Fig. \ref{Fig1}) and \textbf{b} the intuitive approach (see Fig. \ref{Fig2})}
	\label{Fig701}
\end{figure*}

\subsection{Comparison with intuitive approach}
\label{Sect528}

\subsubsection{Classification of fixed-sized image}
\label{Sect581}

For the comparison between the proposed approach (see Fig. \ref{Fig1}) and the intuitive approach (see Fig. \ref{Fig2}), we consider the problem of predicting the permeability of porous media with fixed cubic sizes. Specifically, we consider a size of $48^3$. Similar outputs are observed for other sizes. To ensure a fair comparison, both methodologies are investigated under similar conditions. Specifically, both methods are set to have an approximately equal number of trainable parameters (i.e., 828738 for the intuitive strategy and 828673 for our approach). Accordingly, the size of the vector representing the global feature is 64 in both methods. All other parameters such as the number of modes in each direction, the number of FNO units, the classifier architecture, and size, are the same in both methods and are set as those listed in Sect. \ref{Sect4} (i.e., the training section).

Our results demonstrate that both methods perform proficiently, with the $R^2$ score of 0.99348 and 0.97360 over the test set for the intuitive approach (see Fig. \ref{Fig2}) and the proposed approach (see Fig. \ref{Fig1}), respectively. The evolution of the loss function for the training and validation sets indicates a convergence after approximately 3000 epochs. This deep learning experiment confirms an approximately equivalent computational cost between the two approaches. Hence, when the image size of training data is fixed, both strategies are effective for the defined image classification task and there is no significant advantage for one method over the other, according to our analysis. As a last point in this subsection, we note that one may also use static max pooling in the architecture of the traditional approach since the size of porous media is fixed in this experiment. Based on our results, the performance does not change.

\subsubsection{Classification of multi-sized images}

In this subsection, we compare the performance of the proposed approach (see Fig. \ref{Fig1}) with the intuitive approach (see Fig. \ref{Fig2}) in predicting the permeability of porous media with varying sizes. redFor a fair comparison between the intuitive approach and the proposed approach, we use the same training, validation, and test set described in Sect. \ref{Sect3}. The evolution of both training and validation losses is depicted in Fig. \ref{Fig701}. Figure \ref{Fig701} indicates a divergence between the training and validation losses for the network used in the intuitive approach, which suffers from overfitting, whereas this is not the case for the proposed approach. The superiority of the proposed approach is also evident by the $R^2$ score obtained for the test set. Accordingly, the $R^2$ scores of the proposed approach and the intuitive approach are respectively 0.96809 and $-0.42632$. The negative value of the $R^2$ score for the intuitive approach demonstrates that its model makes worse predictions than a model that simply predicts all outputs as the mean value of the dataset. Note that changing hyper-parameters, such as the number of modes, channel width, and number of FNO layers, does not improve the model of the intuitive approach.

This flaw stems from two reasons. First, using the intuitive approach, the network captures the global feature after lifting cubes into the original space, while the trainable parameters of the network are mainly defined in the Fourier space. Second, the adaptive max pooling's size is altered depending on the size of the input cubic porous medium. These two together lead to a misrepresentation of the global feature of cubes with different sizes, when the network tends to predict the permeability of the validation and test sets. Note that in Sect. \ref{Sect581}, we showed that the intuitive approach worked well when it was trained over porous media with fixed sizes. However, the result of our machine learning experiments illustrates that the global features of cubes with different sizes are amalgamated. In contrast, our approach uses static max pooling consistent with the channel width of Fourier neural operators before transitioning back to the original space. This approach enables the capture of global features prior to changing spaces.

\section{Summary and future outlooks}
\label{Sect6}

In this research study, we introduced a novel deep learning framework based on Fourier neural operators for classifying images with different sizes (see Fig. \ref{Fig1}). Because Fourier neural operators are resolution invariant, they have the potential to be used for the task of multi-sized image classification. To reach this goal, Fourier neural operators must be connected to a classifier, ideally using a pooling operator. To this end, we proposed the novel idea of implementing a static max pooling operator, which functions in a high dimensional space with the size of Fourier channel width. We showed the efficiency and robustness of this framework by predicting the permeability of three-dimensional digital porous media with three different sizes of $40^3$, $48^3$, and $56^3$. We explored the effect of key parameters such as the number of Fourier modes in each dimension, the channel width of the discrete Fourier space, activation functions in different layers, and the number of Fourier units. Additionally, we showed that while the network was only trained on the porous media with the sizes of $40^3$, $48^3$, and $56^3$, it could successfully predict the permeability of the porous media with the sizes of $36^3$, $44^3$, $52^3$, and $60^3$, indicating its generalizability. Moreover, we demonstrated that the idea of implementing an adaptive max pooling (see Fig. \ref{Fig2}), as an intuitive approach for connecting the FNO layers to the classifier, showed a lack of performance when predicting the permeability of porous media of varying sizes. Note that the adaptive max pooling operated in spatial spaces and that pooling had to be adaptive to handle input images with varying sizes.

As a future research direction, we aim to adapt the current architecture and extend its capabilities to image classification. In contrast to the problem of permeability prediction, this approach reduces the problem's dimensionality to two. Additionally, given that the standard dataset for image classification is usually large, we anticipate improved generalizability of the proposed framework. As another research direction, we would like to examine our deep learning framework using real data rather than synthetic data. We aim to expand our work to a variety of porous media, including biological tissues and fuel cells.

\section*{Data availability}
The Python code for the three-dimensional problems is available on the following GitHub repository, \url{https://github.com/Ali-Stanford/FNOMultiSizedImages}.



\section*{Acknowledgements}
Financial support by the Shell-Stanford collaborative project on digital rock physics is acknowledged. Additionally, the first author would like to thank Prof. Gege Wen at Imperial College London for her helpful guidance and discussion on the software engineering aspects of this study. Furthermore, we are grateful to the reviewers for their valuable feedback.







\bibliographystyle{elsarticle-harv} 
\bibliography{example}






\end{document}